\documentclass[10pt,journal]{IEEEtai}
\usepackage{fancyhdr}
\usepackage{extramarks}

\usepackage{amsthm}
\usepackage{amsmath}
\usepackage{amsfonts}
\usepackage{comment}
\usepackage{csquotes}
\usepackage{xspace}
\usepackage{siunitx}

\usepackage{algorithm}

\usepackage{color,array}

\usepackage{amsmath} 
\usepackage{tikz}
\DeclareMathOperator*{\argmin}{arg\,min}
\DeclareMathOperator*{\argmax}{arg\,max}
\usepackage{float}

\usepackage[ruled,algo2e]{algorithm2e}
\usepackage{amsfonts}
\usepackage{amssymb}
\usepackage{algpseudocode}
\usepackage[all]{xy}
\usepackage{tikz-cd}
\usepackage{multirow}
\usepackage{bm}
\usepackage{booktabs}
\usepackage[numbers,sort&compress]{natbib}
\usepackage{graphicx,subfigure}
\usepackage{palatino}
\usepackage{caption}
\usepackage{subcaption}
\usepackage{graphicx}

\usepackage[colorlinks,linkcolor=black,anchorcolor=black,citecolor=black,urlcolor=blue]{hyperref}

\usepackage{booktabs}
\usepackage{mathtools}
\usepackage{amssymb}
\usepackage{capt-of}
\usepackage{mciteplus}
\usepackage{cite}
\usepackage{mathrsfs}
\usepackage{parskip}
\usepackage{soul}
\usepackage{textcomp}
\usepackage[colaction]{multicol}
\usepackage[switch]{lineno}
\usepackage{lipsum}
\usepackage{etoolbox}
\usepackage{longtable}
\usepackage{array}
\usepackage{tablefootnote}
\usepackage{ragged2e}
\usepackage{lscape}
\newcolumntype{C}[1]{>{\centering\arraybackslash}p{#1}}
\usetikzlibrary{automata,positioning}
\newcommand{\km}[1]
{\textcolor{black}{#1}}

\newcommand{\gokul}[1]
{\textcolor{black}{#1}}

\usetikzlibrary{automata,positioning}

\usepackage{tikz,xcolor,hyperref}
\definecolor{lime}{HTML}{A6CE39}
\DeclareRobustCommand{\orcidicon}{%
	\begin{tikzpicture}
	\draw[lime, fill=lime] (0,0)
	circle [radius=0.16]
	node[white] {{\fontfamily{qag}\selectfont \tiny ID}};
	\draw[white, fill=white] (-0.0625,0.095)
	circle [radius=0.007];
	\end{tikzpicture}
	\hspace{-2mm}
}
\foreach \x in {A, ..., Z}{%
	\expandafter\xdef\csname orcid\x\endcsname{\noexpand\href{https://orcid.org/\csname orcidauthor\x\endcsname}{\noexpand\orcidicon}}
}

\begin{document}
\title{
MALADY: Multiclass Active Learning with Auction Dynamics  on Graphs  
}
    
  \author{Gokul Bhusal$^1$, Kevin Miller$^3$,
 Ekaterina Merkurjev$^{1,2}$\footnote{Corresponding author,
	Email:  merkurje@msu.edu}   \\
$^1$ Department of Mathematics, 
Michigan State University, MI 48824, USA.\\
$^2$ Department of Computational Mathematics, Science and Engineering\\
Michigan State University, MI 48824, USA.\\
$^3$  Department of Mathematics,  Brigham Young University, UT 84602, USA. \\
}




\maketitle

\vspace{-0.5cm}
\begin{abstract}

Active learning enhances the performance of machine learning methods, particularly in low-label rate scenarios, by judiciously selecting a limited number of unlabeled data points for labeling, with the goal of improving the performance of an underlying classifier. In this work, we introduce the Multiclass Active Learning with Auction Dynamics on Graphs (MALADY) algorithm, which leverages an auction dynamics technique on similarity graphs for efficient active learning.
In particular, the proposed algorithm incorporates an active learning loop using as its underlying semi-supervised procedure an efficient and effective similarity graph-based auction method consisting of upper and lower bound auctions that integrate class size constraints. 
In addition, we introduce a novel active learning acquisition function that incorporates the dual variable of the auction algorithm to measure the uncertainty in the classifier to
prioritize queries near the decision boundaries between different classes.
Overall, the proposed method can efficiently obtain accurate results using extremely small labeled sets containing just a few elements per class; this is crucial since labeled data is scarce for many applications. Moreover, the proposed technique can incorporate class size information, which improves accuracy even
further. 
Lastly, using experiments on classification tasks and various data sets, we evaluate the performance of our proposed method and show that it exceeds that of comparison algorithms.
\end{abstract}

\begin{IEEEImpStatement}
Sufficient amounts of labeled data are an essential element for good performance in machine learning tasks.
However, in many practical applications, obtaining sufficient amounts of labeled data can be very expensive and time-consuming. Graph-based semi-supervised methods have shown great promise in low-label rate machine learning. However, one can further enhance the performance of semi-supervised learning by integrating an active learning framework. In this work, we propose an active learning setting that incorporates an auction dynamics technique for the semi-supervised learning problem in a similarity graph-based framework. The experimental results demonstrate a significant improvement in the semi-supervised algorithm performance when integrating our active learning method. A wide range of applications, including healthcare, natural language processing, and remote sensing, can greatly benefit from this work in data-efficient machine learning.
\end{IEEEImpStatement}

\vspace{-0.0cm}
\begin{IEEEkeywords} Active Learning, Auction Dynamics for Semi-Supervised Learning, Uncertainty Sampling

\end{IEEEkeywords}

\section{Introduction}
\label{Introduction}
\begin{figure*}
\vspace{-0.5cm}
    \centering
    \includegraphics[width=0.86\linewidth]{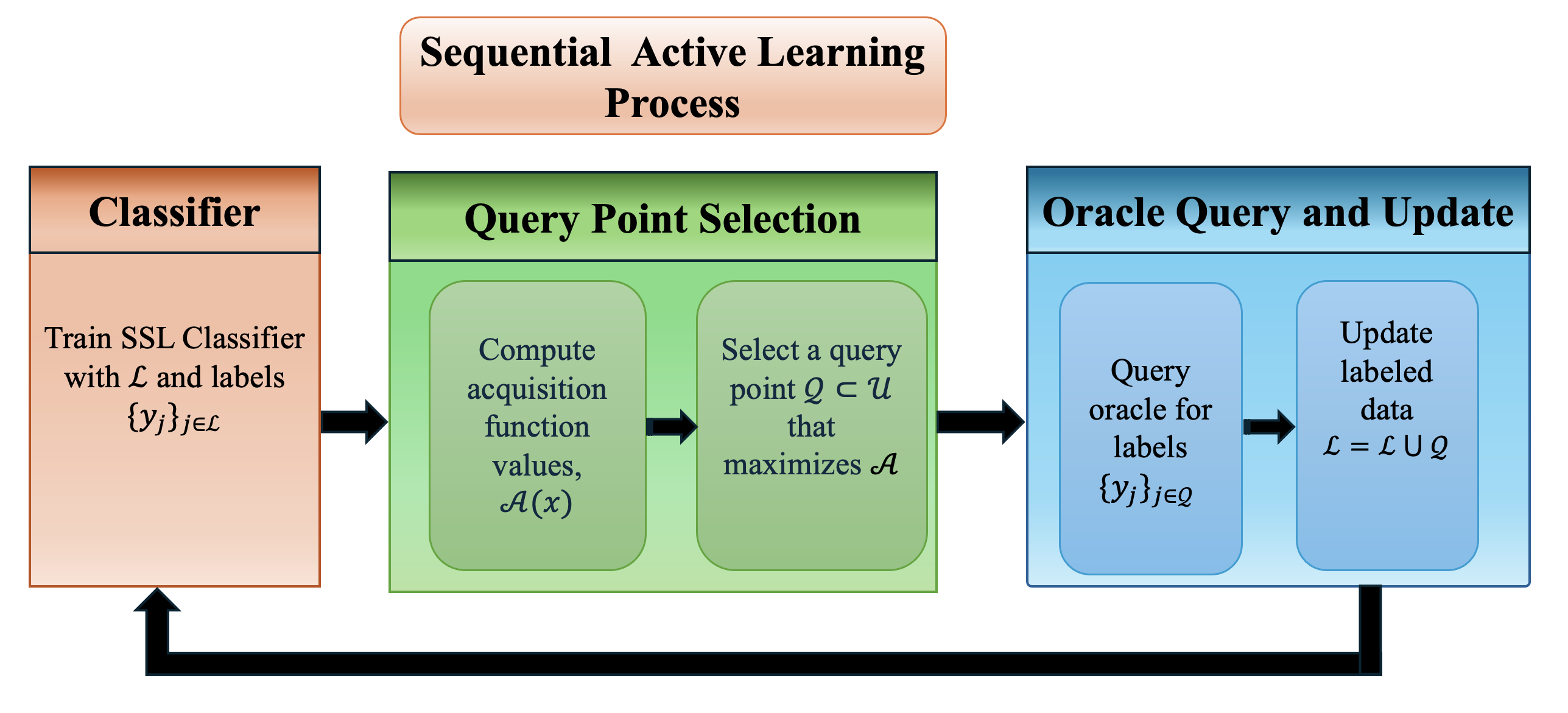}
    \caption{The flowchart of the sequential active learning process. The process starts by training the underlying classifier using the initial labeled set $\mathcal{L}$. A query point $\mathcal{Q}\in\mathcal{U}$ is then selected based on the current acquisition function values. This query point is also labeled according to an oracle (human in the loop or a domain expert), and subsequently, the point is added to the current labeled data $\mathcal{L}$. This process continues until the desired number of points is reached in the labeled set $\mathcal{L}$.  }
    \label{fig:flowchart}
\end{figure*}

\IEEEPARstart{T}{he} choice of training points can have a significant impact on the performance of a machine learning (ML) model, particularly in semi-supervised learning (SSL) scenarios where the labeled set is small. Active learning (AL) is a sub-field of ML that improves the performance of underlying ML methods by carefully selecting unlabeled points to be labeled via the use of a human in the loop or a domain expert. Most AL methods alternate between (1) training a model using current labeled information and (2) selecting query points from an unlabeled set using an acquisition function that quantifies the utility of each point in the unlabeled set. This iterative process of training a classifier and labeling chosen query points is referred to as the AL process, whose flowchart is shown in Figure \ref{fig:flowchart}. Given a dataset $\mathcal{X} = \{x_1, x_2, \ldots, x_N\} \subset \mathbb{R}^d$, we define the \textit{labeled set} $\mathcal{L} \subset \mathcal{X}$ as the set of labeled points with the corresponding labels $y_i = y(x_i) \in \{1, 2, \ldots, C\}$; then, by \textit{labeled data}, we refer to both the labeled set $\mathcal{L}$ of inputs along with their corresponding labels $\{y_i\}_{x_i \in \mathcal{L}}$. The \textit{unlabeled set} then is the set of unlabeled inputs, $\mathcal{U} = \mathcal{X} \setminus \mathcal{L}$. A \textit{query set} $\mathcal{Q} \subset \mathcal{U}$ is the set of unlabeled points that have been chosen to be labeled and added to the labeled set for the next iteration of the AL process.
While the sets $\mathcal{L}$ and $\mathcal{U}$ change throughout the active learning process as query points are labeled, we simplify our notation of these sets to not explicitly denote the iteration; rather, $\mathcal{L},\ \mathcal{U},$ and $\mathcal{Q}$ will respectively denote the labeled, unlabeled, and query sets at the \textit{current} iteration of the AL process.

Given the current $\mathcal{L}$ and $\mathcal{U}$, the main challenge in AL is to design an \textit{acquisition function} $\mathcal{A}: \mathcal{U} \rightarrow \mathbb{R}$ that quantifies the benefit of obtaining the label for each currently unlabeled point. 
The acquisition function values $\mathcal{A}(x)$ for each $x \in \mathcal{U}$ then allow one to prioritize which currently unlabeled data points are chosen to be in the query set $\mathcal{Q}$ to in turn be labeled by a human in the loop or a domain expert. 
When $|\mathcal{Q}| = 1$, the process is referred to as  \textit{sequential} AL while $|\mathcal{Q}| > 1$ corresponds to \textit{batch} AL. In this work, we focus on sequential AL, and Figure \ref{fig:flowchart} reflects this specialization.
While there are various ways to select $\mathcal{Q} \subset \mathcal{U}$ from the set of acquisition function values $\{\mathcal{A}(x) \}_{x \in \mathcal{U}}$, it is most common in sequential AL to use $\mathcal{Q} = \{x^\ast\}$, where $x^\ast = \argmax_{x \in \mathcal{U}} \mathcal{A}(x)$ is the maximizer of the acquisition function. 


Uncertainty sampling-based acquisition functions \citep{settles_active_2012,bertozzi2018uncertainty,qiao_uncertainty_2019} are popular for practical active learning and are frequently efficient to compute in applications. In particular,  uncertainty-based acquisition functions favor query points that are near the current classifier's decision boundary whose inferred classifications are interpreted as most "uncertain". These methods are emblematic of \textit{exploitative} AL since they explicitly use measures of distances to decision boundaries to select query points \citep{miller2023poisson}. Moreover, some of the uncertainty-based acquisition functions designed for graph-based classifiers include the uncertainty norm for Poisson-reweighted Laplace Learning \citep{miller2023poisson}, variance minimization (VOpt) \citep{ji_variance_2012}, model-change (MC) \citep{miller_model-change_2021, miller2022dissertation},
model-change variance optimality (MCVOpt) \citep{miller_model-change_2021, miller2022dissertation}, $\Sigma$-optimality \citep{ma_sigma_2013}, hierarchical sampling for active learning \citep{dasgupta_hierarchical_2008}, cautious active clustering \citep{cloninger_cautious_2021}, and shortest-shortest path $S^2$ \citep{dasarathy_s2_2015}. In this paper, we introduce a novel uncertainty-based acquisition function that leverages the computed optimal price and incentive values from the underlying auction dynamics graph-based classifier. 

Acquisition functions for the AL process have also been integrated into deep learning frameworks to boost the performance of such parametric models. In particular, some of the recent works include batch AL by diverse gradient embeddings (BADGE) \citep{ash2020badge}, AL framework in Bayesian deep learning \citep{gal_deep_2017}, diverse batch acquisition for deep Bayesian AL \citep{kirsch2019batchbald}, diffusion-based deep AL \citep{kushnir2020diffusion}, stochastic batch acquisition functions \citep{kirsch2021stochastic}, and AL for convolutional neural networks \citep{sener_active_2018}.

On the other hand, an important component of the AL process is the underlying method used for training a classifier based on the currently labeled data $\mathcal{L}$. To this end, (similarity) graph-based methods have proven to be effective models for semi-supervised and AL settings, particularly in the low label rate regime \citep{song2022graph, calder2020poisson, calder2023rates, bhusal2024persistent}. These methods leverage both the labeled and unlabeled sets to construct a similarity graph on which the observed labels on $\mathcal{L}$ are used to make inferences of the labels on the unlabeled data according to the graph topology; thus, the geometric structure of the dataset is modeled with the graph and similar data points receive similar inferred outputs. Such techniques are beneficial since they utilize the structure of and the information from the usually abundant unlabeled data to build a more effective model.

In contrast, variants of auction algorithms, originally developed by Bertsekas \citep{bertsekas1979distributed, bertsekas1991linear, bertsekas1989auction,bertsekas1998network} for solving the classic assignment problem, have a simple intuitive structure, are easy to code, and have excellent performance in both accuracy and timing. 
In fact, auction methods substantially outperform their main competitors for important problems, both in theory and in practice, and are also naturally well-suited for parallel computation during coding. Therefore, in this paper, we integrate (similarity) graph-based techniques and auction-based procedures to derive the underlying algorithm used in the active learning process, as well as the acquisition function. 

In \citep{jacobs2018auction}, the authors have shown how auction algorithms can be utilized for various applications, such as semi-supervised graph-based learning and classification in the presence of equality or inequality class size constraints on the individual classes. Modifications of this technique for the classification of 3D sensory data are discussed in \citep{merkurjev2020fast}. Moreover, auction methods can be used in Merriman, Bence, and Osher (MBO)-based threshold dynamics schemes, which involve threshold dynamics to tackle different tasks; this was shown in \citep{jacobs2018auction}. For example, the original MBO scheme \citep{merriman1992diffusion} was developed in 1992 to simulate and approximate motion by mean curvature; the procedure consists of alternating between diffusion and thresholding. Such an MBO technique was then modified and adapted to a similarity graph setting for the purpose of binary classification and image inpainting in \citep{merkurjev2013mbo}, with a multiclass extension developed in \citep{garcia2014multiclass} and \citep{merkurjev2014diffuse}. Applications of the graph-based technique to heat kernel PageRank and hyperspectral imagery and video are detailed in \citep{merkurjev2018semi} and \citep{merkurjev2014graph}, respectively. A summary of recent graph-based optimization approaches for machine learning, including MBO-based ones, is presented in \citep{bertozzi2019graph}. 

\km{Moreover, graph neural networks (GNN) \citep{wu2020comprehensive} constitute an important area of modern, graph-based learning methods that have found application in node classification, graph classification, and link prediction. These methods leverage a graph's structure of nodes and edges to define compositions of transformations of node features in a fashion that is similar to deep neural networks. Notable contributions in this area include \citep{kipf2016semi, velickovic2017graph, li2024permutation}, with advances in graph classification highlighted in \citep{bai2024haqjsk, han2022g}.}

With regards to active learning for graph-based semi-supervised classifiers, various recent pipelines have been developed for image processing applications, e.g., image segmentation \citep{chen2024batch}, surface water and sediment detection \citep{chen2023graph}, classification of synthetic aperture radar (SAR) data \citep{miller2022graph, chapman2023batchALsar}, unsupervised clustering of hyperspectral images using nonlinear diffusion \citep{murphy_unsupervised_2019}, hyperspectral unmixing \citep{chen2023graph}, medical image analysis \citep{budd2021survey}, hyperspectral image classification \citep{cao2020hyperspectral}, and subgraph matching \citep{ge2025iterative}.


Overall, in this paper, we integrate active learning components to develop MALADY, an accurate active learning algorithm incorporating auction dynamics on similarity graphs for semi-supervised learning tasks.

The contributions of the paper are as follows:

\begin{itemize}

\item We propose a new active learning technique based on auction dynamics. For the underlying semi-supervised similarity graph-based classifier in our active learning process, we incorporate elements of the auction dynamics algorithm for semi-supervised learning in \citep{jacobs2018auction} and use a more general optimization energy functional. The general formulation can be reformulated as a series of modified assignment problems, each of which can be tackled using auction methods incorporating class size constraints. 
\vspace{0.2cm}
\item We propose a novel acquisition function for the active learning process that uses the dual variables of the dual formulations of the modified assignment problems incorporating upper and lower bound class size constraints. The proposed acquisition function not only measures the uncertainty of the volume bound auction algorithm for graph-based semi-supervised problems but also captures salient geometric classifier information that can be leveraged for exploitation of decision boundaries. 
\vspace{0.2cm}
 \item  The proposed method is able to perform accurately even when the labeled set is extremely small, as small as just a few labeled elements per class. This is important since labeled data is scarce for many applications. In addition, the algorithm integrates class size information, improving accuracy further.
 \vspace{0.2cm}
\item We conduct experiments on data classification using various datasets, and the results demonstrate that our proposed framework performs more accurately compared to other state-of-the-art methods.
\end{itemize}
The remainder of the paper is organized as follows: in Section \ref{Background}, we present background information on the graph-based framework and the auction algorithm with volume constraints. In Section \ref{Method}, we derive our semi-supervised classifier and the proposed acquisition function.
The results of the experiments on benchmark data sets and the discussion of the results are presented in Section \ref{Results}. Section \ref{Conclusion} provides concluding remarks.

\section{Background} \label{Background}
In this background section, we first discuss the (similarity) graph construction technique, which is fundamental for graph-based methods. Then, we discuss auction dynamics with volume constraints for semi-supervised learning \citep{jacobs2018auction} as the technique that is fundamental for the construction of our proposed semi-supervised method. The last subsection reviews some semi-supervised learning algorithms using auction dynamics procedures. \gokul{The notation used in this paper is described in Table \ref{tab:notations}. }
\begin{table}[!ht]
    \centering
    \vspace{0.05cm}
    \begin{tabular}{lc}
        \hline
        Variable & Description\\
        \hline
        $N$ & number of nodes\\
        $\mathcal{X}$ & dataset, vertex set\\
        $\mathcal{L}$ & labeled set\\
        $\mathcal{U}$ & unlabeled set\\
        $\mathcal{Q}$ & query set\\
        $K$ & number of similarity classes\\
        $U_i$ & upper bound for class $i$\\
        $B_i$ & lower bound for class $i$\\
        $S_i$ &similarity class for class $i$\\
        $V_i$ & cardinality of similarity class $S_i$\\
        \hline
    \end{tabular}
    \caption{Notation Table}
    \label{tab:notations}
\end{table}

\vspace{-0.2cm}

\subsection{Graph Construction}
\label{Graph_Construction}
In this section, we review the graph-based framework we use in this paper. Consider a dataset $\mathcal{X} = \{x_1 , x_2, \cdots x_N\} \in \mathbb{R}^d$, where each data element is represented by a $d$-dimensional feature vector. We generate a undirected graph $G(\mathcal{X}, W)$ of vertices and edges between vertices, where $\mathcal{X}$ is set of vertices representing the data elements, and $W \in \mathbb{R}^{N \times N}$ is the weight matrix, consisting of weights between pairs of vertices. The weight matrix $W$ is computed using a weight function $w: \mathcal{X} \times \mathcal{X} \rightarrow \mathbb{R}$, where $W_{i,j}= w(x_i,x_j)$ denotes the weight on the edge between the vertices $x_i$ and  $x_j$, where $i \neq j$. Overall, the weight matrix quantifies the similarities between the feature vectors, i.e., data elements, whether labeled or unlabeled; thus, the graph-based framework can provide crucial information about the data and leverage information from the abundant unlabeled data. Some popular weight functions include: 
\begin{itemize}
\item Gaussian weight function
\begin{equation}
\label{gaussian}
w(x_i,x_j) = \exp\left(-\frac{d(x_{i},x_{j})^2}{\sigma ^2}\right),
\end{equation}
where $d(x_{i},x_{j})$ represents a distance (computed using a chosen measure, such as Euclidean distance) between vertices $x_{i}$ and $x_{j}$, associated with the $i^{th}$ and $j^{th}$ data elements, and $\sigma\hspace{-0.05cm}>\hspace{-0.05cm}0$ is a parameter which controls the scaling in the weight function.
\vspace{0.2cm}
\item cosine similarity weight function \citep{singhal2001modern}
\begin{equation}
\label{cosine}
w(x_i,x_j) = \exp\left(-\frac{d(x_{i},x_{j})^2}{\sigma(x) \sigma(y)}\right),
\end{equation}
where $d(x_i , x_j) = arccos\left(\frac{\langle x_i , x_j\rangle}{||x_i||||x_j||} \right)$ and $\sqrt{\sigma(x_i)} = d(x_i , x_M)$ is a local parameter for each $x_i$, and $x_M$ is the $M^{th}$ closest vector to $x_i$.
\end{itemize}

With the assumption that the high-dimensional data is concentrated near a low-dimensional manifold and the manifold is locally Euclidean, we only compute the $k$ nearest neighbors (kNN) of each point using an approximate nearest neighbor search algorithm.  This ensures that the weight matrix is sparse and enhances efficiency.  For an approximate nearest neighbor search, we use the GraphLearning package \citep{calder2020poisson}, which uses the Annoy library \citep{annoy}. To preserve the symmetric property of the weight matrix, we calculate ${W^{final}}_{ij} = \{W_{ij} + W_{ji}\}/2$.

\vspace{-0.0cm} 

\subsection{Auction Algorithm with Exact Volume Constraints}
Since our proposed method involves auction techniques originally designed to solve the assignment problem, we review the procedures here. Given two disjoint sets $\mathcal{X}$ and $\mathcal{Y}$ of same cardinality $r$ and a benefit function $c: \mathcal{X} \times \mathcal{Y} \rightarrow\mathbb{R}$, the assignment problem aims to identify a one-to-one correspondence $ M = \{(x_1 ,y_1),\cdots, (x_r,y_r)\} $ of sets $X$ and $Y$, so that the total benefit of the matching 
\begin{equation}
\label{assignment_problem}
    \sum_{(x,y)\in M} c(x, y)
\end{equation}
is maximized. 
The assignment problem (\ref{assignment_problem}) can be restated as the following optimization problem by representing the matching as a binary vector $z$, where $z_{y}(x) = 1$ if $(x,y)$ are matched and $0$ otherwise:
\begin{equation}
\begin{split}
\label{eq:assignment01} & \max_{\mathbf{z}:\mathcal{X}\times \mathcal{Y}\to\{0,1\}} \sum_{x\in \mathcal{X}}\sum_{y \in \mathcal{Y}} c(x,y)z_{y}(x) \\& \quad \textrm{s.t.} \; \sum_{x\in \mathcal{X}} z_y(x)=1 \; \; \; \forall y , \; \sum_{y\in \mathcal{Y}} z_y(x)=1 \; \; \; \forall x.
\vspace{-0.5cm}
\end{split}
\end{equation}
Moreover, the above optimization problem can be written as a classical linear programming problem ($LP$) if we relax the binary constraint on $\bm{z}$: 
\begin{equation}
\begin{split}
\label{LP}
\begin{aligned}
& \max_{\bm{z \ge 0}}  \sum_{x \in \mathcal{X}} \sum_{y \in \mathcal{Y}} c(x,y)z_{y}(x) \hspace{0.5cm}
\\&  \textrm{s.t.}  \sum_{x \in \mathcal{X}} z_{y}(x) =1,
   \hspace{0.3cm} \sum_{y \in \mathcal{Y}} z_{y}(x) =1. \hspace{0.5cm} (LP) \\
\end{aligned}
\end{split}
\end{equation}

This relaxation turns out to be exact  
\km{since the solution to a bounded and feasible linear programming problem always includes a vertex of the feasible polytope $\mathcal{P}$. The relaxed constraint set in (\ref{LP}) is the polytope $\mathcal{P} = \left\{ \bm{z} \geq 0 : \sum_{x \in \mathcal{X}} z_y(x) = 1, \sum_{y \in \mathcal{Y}} z_y(x) = 1 \right\} $, whose vertices are precisely vectors $\bm{z}$ whose entries are binary.}
Furthermore, we can split $Y$ into $K$ similarity classes $\{S_i\}_{i=1}^K$ each of size $V_i$, and let $c (x,y) = a_i(x)$ for each $y\in S_{i}$. With these choices, we can reformulate (\ref{LP}): 
\begin{equation}
\begin{split}
\begin{aligned}
& \max_{\bm{z \ge 0}}  \sum_{x \in \mathcal{X}} \sum_{i=1}^K a_{i}(x) \sum_{y \in S_i} z_{y}(x)\hspace{0.5cm} \\&
\textrm{s.t.}   \sum_{x \in \mathcal{X}} z_{y}(x) =1,
   \hspace{0.3cm} \sum_{y \in \mathcal{Y}} z_{y}(x) =1 \hspace{0.5cm}  \\
\end{aligned}
\end{split}
\end{equation}

Using $u_i(x)= \sum_{y \in S_i} z_{y}(x)$, we obtain equivalently:
\vspace{-0.2cm}
\begin{equation}
\begin{split}
\begin{aligned}
& \max_{\bm{u \ge 0}}  \sum_{x \in \mathcal{X}} \sum_{i=1}^K a_{i}(x) u_i(x)\hspace{0.5cm}
\\& \textrm{s.t.}  \sum_{x \in \mathcal{X}} u_i(x) =V_i,
   \hspace{0.3cm} \sum_{i=1}^K u_i(x) =1. \hspace{0.5cm}  \\
\end{aligned}
\label{target_problem}
\end{split}
\end{equation}

We will now focus on solving \eqref{target_problem}, which is a special case of \eqref{LP}. First, it is helpful to practically interpret \eqref{target_problem}. In particular, one can view each class in the data classification problem as an institution, such as a gym, that offers a certain number of memberships and the data elements as people trying to obtain a deal on buying access to only one of the institutions. Moreover, the coefficients $a_i(x)$ can represent person $x$'s desire to buy the membership in institution $i$. In this view, the solution to \eqref{target_problem} maximizes the total satisfaction of the population. Notice that the class size constraints (i.e. number of memberships available) make the assignment of institution memberships to people nontrivial. 

  One approach to assigning memberships involves the market mechanism where an institution $j$ is equipped with a price $p_j$. Then, person $x$ will want to buy the membership from an institution that offers the best value:
\begin{equation}
\label{profit_def}
j^*(x) = \argmax_{1\leq j \leq N} \hspace{0.1cm} a_{j} (x) - p_{j}
\end{equation}

However, the resulting matching will probably not satisfy the class size constraints. Thus, the challenge is to compute an optimal price vector, called an equilibrium price vector, that results in an institution-person matching that satisfies the constraints on the number of available memberships. The answer lies in the dual formulation of the assignment problem, which can be shown to be the following problem:
\begin{equation}
\begin{aligned}
\min_{\bm{p} \in \mathbb{R}^K, \bm{\pi} \in  \mathbb{R}^N} \quad  & \sum_{i = 1}^{K} p_{i}S_i + \sum_{x} \bm{\pi} (x) \\
\textrm{subject to} \quad & \bm{\pi}(x) + p_{i} \geq a_{i}(x).
\end{aligned}
\label{dual}
\end{equation}

In fact, it turns out that the equilibrium price vector $p_*$ of the primal problem is in fact the optimal solution of the dual problem \eqref{dual}. Moreover, the optimal value of $\bm{\pi}$ is completely determined by $\bm{p}$; given a price vector $\bm{p}$, the above dual problem is minimized when $\pi(x)$  equals the maximum value of $a_{i}(x)- p_{i}$ over $i$. 

According to the complementary slackness (CS) condition, a complete assignment $M$ and a price vector $\bm{p}$ are primal and dual optimal if and only if each person is assigned to an institution offering the best deal to them. Moreover, the CS condition can be relaxed to allow a person to be assigned to institutions that are within $\epsilon$ of achieving the best deal for them in the definition \eqref{profit_def}. In particular, we say that an assignment $M$ and a price vector $\bm{p}$ satisfies $\epsilon$-complementary slackness ($\epsilon$-CS) if
\begin{equation}
\pi(x) - \epsilon \leq   \hspace{0.1cm} a_i(x) - p_{i} \; \; \text{for all} \; \; i.
\end{equation}
This modification is very useful in the case of a price war. In a price war, multiple people compete for the same memberships without ever raising the prices of the institutions, trapping the algorithm in an infinite loop.

In \citep{bertsekas1979distributed}, Bertsekas detailed an auction technique for solving the assignment problem by finding the equilibrium (optimal) price vector. Since the seminal paper, Bertsekas and others have extended the technique to more general problems and have improved the computational aspects. For example, \citep{bertsekas1989auction} develops a technique that efficiently handles assignment problems with multiple identical objects. Overall, an exhaustive reference on auction algorithms is included in \citep{bertsekas1991linear}, which contains information on linear network optimization. Continuous and discrete models for network optimization are described in \citep{bertsekas1998network}.

Bertsekas's auction technique contains the following steps: at the beginning of an iteration, one starts with an assignment and price vector satisfying $\epsilon$-CS with $\epsilon > 0$. There are two phases in each iteration: the bidding phase and the assignment phase (the assignment $M$ and price vector $\bm{p}$ is updated while maintaining $\epsilon$-CS): 
\hspace{0.7cm}
\begin{itemize}
    \item Bidding phase: For each $x$, under the assignment $M$:
    \vspace{-0.1cm}
    \begin{itemize}
        \item  Compute the current value $v_i= a_i(x) - p_i$ for each class $i$, and choose $i^* \in \argmax v_i$.
            \vspace{0.05cm}
    \item Find best value offered by a class other than $i^*$: \begin{equation}
         u_i= \argmax_{i \neq i^*} v_i.
    \end{equation}
         \item Compute the bid $b(x)$ of element $x$ for class $i^*$: 
    \begin{equation}
        b(x) = p_{i^*}+ \epsilon + v_{i^*}  -  u_i  
    \end{equation}
    \end{itemize}

\item Assignment phase: 
    \vspace{0.00cm}
\begin{itemize}
    \item 
    If class $i^*$ has already given out $V_{i^*}$ memberships, remove $y$ currently assigned to the class with the lowest bid and add $x$ to class $i^*$, and set $p_{i^*}$ to be the minimum bid value over all currently assigned to class $i^*$.
        \vspace{0.00cm}
    \item If class $i^*$ has not yet given out all $V_{i^*}$ memberships, add $x$ to class $i^*$. If now all memberships of class $i^*$ are bought, set $p_{i^*}$ to be the minimum bid value over all currently assigned to class $i^*$. Otherwise, the price remains the same.
\end{itemize}
\end{itemize}
\vspace{0.2cm}
The Membership Auction algorithm described above is given in Appendix as Algorithm \ref{alg:faso}. 

\subsection{Auction Algorithm with Upper and Lower Bound Volume Constraints}
\label{volume_constraints}
In many scenarios, the exact number of elements in each class is unknown. Thus, in \citep{jacobs2018auction}, the authors present an upper and lower bound auction algorithm for semi-supervised learning that allows the size of each class to fluctuate between upper and lower bounds. In particular, let $K$ be the number of classes, and suppose that class $j$ must have at least $B_j$ members and at most $U_j$ members, where $B_{j} \leq U_{j}$, and $\sum_{1}^K B_j \leq N \leq \sum_{j=1}^K U_j$, where $N$ is the number of data elements in the data set.

With these additional constraints, the modified version of the assignment problem becomes: 
\begin{equation}
\begin{split}
\label{eq:ulassign}
   & \max_{u \geq 0 } \sum_{x} \sum_{i=1}^K  a_{i}(x) u_{i}(x) \hspace{0.5cm}
\\& \textrm{s.t.}   \sum_{i=1}^N u_i(x) =1 , \; B_{i} \leq \sum_{x} u_{i}(x) \leq U_{i} \; \; \forall i.
\end{split}
\end{equation}

The addition of these bounds in the optimization problem introduces some complexities in the problem. 
For example, each data element $x$ always aims to obtain the most desirable class $j$, which may result in a deficiency of members in other classes. To solve this problem, the authors of \citep{jacobs2018auction} introduce an idea of incentives $\bm{t}$ in the market mechanics. In particular, class $i$ must sell $B_i$ memberships, and if it is having trouble attracting a certain number of members, it should offer incentives to attract elements to join the class. This results in competition among the classes, and the classes that are deficient in members are forced to offer competitive incentives to attract the necessary number of data elements. To satisfy the lower bounds, one needs to apply an adaptation of the reverse auction algorithm of Bertsekas and coauthors \citep{bertsekas1993reverse} and have the classes bid on the data elements. 

Following the work of \citep{jacobs2018auction}, the modified dual problem with price and incentives is given as 
\vspace{-0.00cm}
\begin{equation}
\begin{aligned}
\min_{\bm{p}\geq 0, \bm{t}\geq 0, \bm{\pi} \in \mathbb{R}^N} \quad  & \sum_{i = 1}^{K} (p_{i}U_i- t_i B_i) + \sum_{x} \bm{\pi} (x) \\
\textrm{subject to} \quad & \bm{\pi}(x) + p_{i} - t_i \geq a_{i}(x) \; \forall i.
\end{aligned}
\label{eq:uldual}
\end{equation}

The complementary slackness condition for (\ref{eq:ulassign}) and (\ref{eq:uldual}) states that $\bm{u}$ and $(\bm{p},\bm{t}, \bm{\pi})$ are optimal if and only if 
\vspace{-0.2cm} 
\begin{equation}
\begin{split}
\begin{aligned}
 \label{eq:ulcs} 
 &\sum_{i=1}^K\sum_{x} u_{i}(x)(a_{i}(x)-p_{i}+t_{i}-\pi(x))\\&+\sum_{i=1}^K p_{i}(U_{i}-\sum_{x} u_{i}(x))  +\sum_{i=1}^K t_{i}(\sum_{x} u_{i}(x)-B_{i})=0.
\end{aligned}
\end{split}
\end{equation}

\vspace{-0.4cm} 
With the addition of incentives, the  $\epsilon$-CS condition for every matched pair $(x,i)$ satisfies 
\begin{equation}
   a_{i}(x) - p_{i} + t_{i} + \epsilon \geq \max_{1 \leq j \leq N} a_{j}(x_i) - p_{j} +t_j. 
   \end{equation}
    Unfortunately, the last two terms in (\ref{eq:ulcs}) do not have useful $\epsilon$ relaxations. Thus, \citep{jacobs2018auction} proposes a two-stage method to solve (\ref{eq:ulassign}). First, one should run the forward auction algorithm, Algorithm \ref{alg:fasotr} in the Appendix, to satisfy the upper bound constraints with a complete $\epsilon$-CS matching. The output of this stage is then fed into a lower bound auction, Algorithm \ref{alg:rasopr} in the Appendix; the goal of the lower bound auction is to obtain a matching that satisfies the lower bound class size constraints as well. Thus, at the end, one will obtain a $\epsilon$-CS matching satisfying {\it both} the upper and lower bound class constraints. We refer the reader to \citep{jacobs2018auction} for a detailed description. 

\section{Proposed Method} \label{Method}
\subsection{Proposed Semi-Supervised Learning Framework} \label{ssl}
\subsubsection{Notation}
We use a graph-based framework to derive our underlying semi-supervised classifier. Let $K$ denote the number of classes, and let the data set $\mathcal{X}$ consist of the training elements $\mathcal{L}$ with label information and $\mathcal{U}$, the unlabeled training data elements. We embed our data set into a weighted similarity graph $G(\mathcal{X}
,W)$ using the similarity functions mentioned in Section \ref{Graph_Construction}. For class $i$, let $B_i$ and $U_i$ denote the lower bounds and upper bounds of the class, respectively. If exact class sizes are known, we can set $B_i = U_i$, and when the class information is not available, we simply set $B_i =0$ and $U_i = |\mathcal{X}|$. Let $l_x \in \{1,...,K\}$ denote the label of $x \in \mathcal{L}$, $e_i$ denote a vector with $1$ in the $i^{th}$ place and $0$ elsewhere.

\vspace{0.15cm}

\subsubsection{Derivation}
The proposed method is derived using constrained optimization of an energy of the form
\begin{equation}
\label{energy}
    E(\bm{\mathcal{X}}) = R(\bm{\mathcal{X}}) + F (\bm{\mathcal{X}}),
\end{equation}
where $R(\bm{\mathcal{X}})$ is a regularizing term which ensures the smoothness in the partition $\bm{\mathcal{X}}= ({\mathcal{X}}_1, ... , {\mathcal{X}}_K)$ into $K$ classes, and $F(\bm{\mathcal{X}})$ is a fidelity term containing information from the labeled training data. We can reformulate (\ref{energy}) by imposing constraints that incorporate the labeled training data and class size information:
\begin{equation}
\small
    B_i \leq |\mathcal{X}_i| \leq U_i, \quad \mathcal{L}_i \subset \mathcal{X}_i \quad \text{for all} \quad  1 \leq i \leq K,
\end{equation}
where $U_i$ and $B_i$ are upper and lower bounds on the class sizes, respectively, and  $\mathcal{L}_i$ is the set of labeled training points with label $i$. Then, to guarantee a notion of smoothness for partitions of the data set $\mathcal{X}$, one can use a weighted graph cut as a regularizing term. The weighted graph cut is defined as the following term:
 \begin{equation}
     \text{Cut}(\bm{\mathcal{X}}) =  \frac{1}{2} \sum_{i=1}^K \sum_{x\in \mathcal{X}_i} \sum_{y \notin \mathcal{X}_i} w(x,y),
 \end{equation}
where the entries $w(x,y)$ describe how strongly the points $x$ and $y$ are connected. Unfortunately, the graph cut problem is NP-hard. Instead, similarly to \citep{jacobs2018auction, esedoglu2017convolution}, one can consider the graph heat content (GHC) as a convex relaxation of the graph cut. It is defined as:
\begin{equation}
\small
\label{graphheatcontent}
    \text{GHC}(\bm{u}, W) = \sum_{i=1}^K \sum_{x,y \in \mathcal{X}} w(x,y) u_{i}(x) (1-u_i (y)),
\end{equation}
where $\bm{u}: \mathcal{X} \to  \mathbb{R}^K$ is an element of the convex relaxation of the space of $K$-phase partitions of $\mathcal{X}$. For each $x \in \mathcal{X},u(x)= (u_1(x), \cdots, u_K(x)) \in \mathbb{R}^K$, where $u_i(x)$ is the $i^{th} $ entry of vector $u(x)$.  As long as the matrix $W$, for which $W_{ij}=w(x_i,x_j)$, is a positive semi-definite matrix, the above GHC term is a concave term.

With this in mind, we propose the following problem:
\begin{equation}
\begin{split}
\label{eq:unsupervised}
    &\min_{\bm{u}} \; \; \text{GHC}(\bm{u},W) + \text{J}_{conc}(\bm{u})  \\& \text{subject to} \; \; \sum_{i=1}^K u_i(x) = 1 \; \; \; \forall x,\\& u(x) = e_{l_x} \; \; \; \; \forall x \in \mathcal{L}, \\& B_i \leq \sum_{x \in \mathcal{U}} u_i(x) \leq U_i \; \; \; \; \forall i
    \end{split}
\end{equation}
over ${\bm{u}:\mathcal{X} \rightarrow \mathbb{R}^K,\bm{u}\geq 0}$. In this formulation, $\text{GHC}(\bm{u},W)$ is the graph heat content energy in \eqref{graphheatcontent}, which depends on the graph weights, and $\text{J}_{\text{conc}}(\bm{u})$ is any concave function of $u$, including any linear function in $\bm{u}$.  

The $\text{J}_{\text{conc}}(\bm{u})$ term is a term that allows one to consider a wide range of optimization problems within our framework. Without this term, the problem would only involve minimizing the graph heat content subject to certain constraints. The $\text{J}_{\text{conc}}(\bm{u})$ term and some examples of it one can use are described at the end of Section IIIA.
 
The motivation for considering the model \eqref{eq:unsupervised} is: 
\begin{itemize}
\itemsep0em 

\item The first term of \eqref{eq:unsupervised} is a convexified graph cut, and its minimizer groups data elements so that those of different classes are as dissimilar as possible.
\vspace{0.1cm}
\item One can integrate information about class sizes, which can often improve algorithm performance.
\vspace{0.1cm}
\item It enables one to incorporate a combination of weighted edge-based, class-based, and label-based terms that contain important information about the data set elements. This can improve accuracy.
\vspace{0.1cm}
\item For a positive semi-definite matrix $W_{ij}=w(x_i,x_j)$, the energy is a concave function of $u$. Therefore, one may obtain a minimization procedure for \eqref{eq:unsupervised} by considering a "gradient flow" scheme involving linearizations of the energy under constraints.
\end{itemize}
\vspace{0.0cm}
Overall, each step of the scheme to solve \eqref{eq:unsupervised} is equivalent to finding the ${(k+1)}^{th}$ iterate $\bm{u}^{k+1}$ as a solution of the following modified assignment problem:
\vspace{-0.185cm}

 \vspace{-0.2cm}
 
\small
  
 \begin{equation}
  \label{eq:ulassign2} 
 \small
\argmax_{\bm{u}:\mathcal{X} \rightarrow \mathbb{R}^K,\bm{u}\geq 0} \sum_{i=1}^K \sum_{x\in \mathcal{U}}(1- {(\nabla \text{J}_{\text{conc}})}_i(x)   - \sum_{y\notin \mathcal{X}_i^k} w(x,y))u_{i}(x)
 \end{equation}

\vspace{-0.8cm}

 \begin{equation}
 \begin{split}
 \label{part2}
&  \textrm{s.t.} \small \; \sum_{i=1}^K u_{i}(x)=1, u(x)=e_{l_x} \; \;\; \forall x \in \mathcal{L}, \; \\&L_i \leq \sum_{x\in \mathcal{U}} u_i(x) \leq U_i \quad \; \forall i,
   \vspace{-0.125cm}
   \end{split}
\end{equation}
\normalsize
\vspace{0.2cm}
At each iteration, a partition of the data can be formed via $\mathcal{X}_i^{k+1}= \{x  \hspace{0.2cm}  \textrm{s.t.} \hspace{0.2cm} \argmax_{j} u_j^{k+1}(x)=i\}.$
Moreover, the elements $\{a_i(x)\}$, where
\begin{equation}
\label{aucion_coefficient}
   a_i(x) =  (1- {(\nabla \text{J}_{\text{conc}})}_i(x) -\sum_{y\notin \mathcal{X}_i^k} w(x,y)),
\end{equation}
are referred to as coefficients of the modified assignment problem. Note that the above scheme is equivalent to the upper and lower bound assignment problem (\ref{eq:ulassign}).  Our goal is to solve each step of the scheme in \eqref{eq:ulassign2}-\eqref{part2} with an auction-type algorithm that incorporates lower and upper bound class size constraints and also integrates a novel active learning component which will be described in future sections of the paper. 

One can solve each step of the scheme in \eqref{eq:ulassign2}-\eqref{part2} using an auction-type technique similar to the one used in \citep{jacobs2018auction} and described in Section IIC. In particular, one alternates between an upper bound auction procedure and a lower bound auction procedure while decreasing the $\epsilon$ parameter after each lower bound auction. This $\epsilon-$scaling procedure is used to reduce the complexity of the technique and is described in more detail in \citep{jacobs2018auction}. Our semi-supervised learning framework to be used in the active learning process is described as Algorithm 1.

In general, one can consider many choices for the function $\text{J}_{\text{conc}}$. For example, in this paper, we consider a Poisson term, such as the one defined in \citep{calder2020poisson}, as our concave function $\text{J}_{\text{conc}}$. In particular, we consider $\text{J}_{\text{conc}} = - \gamma  \sum_{j=1}^{k} (l_j-\bar{l}) \cdot u(x_j)$, where $k$ is the number of labeled elements, $\{x_1, ..., x_k\}$ are the labeled data elements with labels $\{l_1,..., l_k\} \hspace{-0.05cm}  \in \hspace{-0.05cm}  \{e_1, ..., e_k\}$, which are indicator vectors in $\mathbb{R}^k$, $\gamma>0$, and $\bar{l}= \frac{1}{k}\sum_{j=1}^k l_j$. An optimization problem consisting of this term, coupled with a graph-based regularizer \eqref{graphheatcontent}, will result in a method that is provably advantageous for low label rates; please refer to \citep{calder2020poisson} for a theoretical justification of Poisson-like terms. 

Another example of $\text{J}_{\text{conc}}$ one can use is the modularity term in \citep{hu2013method}\hspace{0.05cm}: $-\gamma \sum_{i} \sum_{x \in \mathcal{X}} k_x \lvert u_i(x)-\frac{1}{2r}\sum_{y \in V}k_y u_i(y) \rvert^2$, where a total variation optimization problem with this term is shown to be equivalent to modularity optimization, which tries to detect communities in the graph based on their modularity. Here, $k_x= \sum_y w(x,y)$ is the degree of vertex $x$, $\gamma>0$ is a resolution parameter, and $2r= \sum_x k_x$. Using this term will open new avenues to approach network science and detection of communities.

Lastly, it is clear that any linear term in $u$, such as $\sum_{i=1}^{m} \sum_{x \in \mathcal{X}} R_i(x)u_i(x)$, can be used for $\text{J}_{\text{conc}}$. In particular, one can assign $R_i(x)$ to be a class homogeneity term, where $R_i(x)$ is the cost of assigning $x$ to class $i$. In the latter case, one should formulate $R_i(x)$ so that it is small if $x$ is likely to belong to class $i$ and large otherwise. The terms may be defined using the eigenvectors of the graph Laplacian or using a fit to an expected value of a variable. Please refer to \citep{bae2017convex} for a more detailed description of class homogeneity terms.

\vspace{0.15cm}

\begin{algorithm2e}[h]
\SetAlgoLined
\SetKwInOut{Input}{Input}
\Input{Vertex set $\mathcal{X}$, Weight matrix $W$, initial partition $\bm{\mathcal{X}}$, lower bound $\bm{B}$, upper bound $\bm{U}$, time step $\delta t$, number of steps $s$, auction error tolerance $\epsilon_{min}$, epsilon scaling factor $\alpha$, initial epsilon $\epsilon_0$. }
\KwResult{Final configuration $\bm{\mathcal{X}}^{s}$, price vector $\bm{p}$, incentives vector $\bm{t}$}
 \textbf{Initialization}: Set $\bm{\mathcal{X}}^0:=\bm{\mathcal{X}}$, set $\bar{\epsilon}=\epsilon_{min}/N$\;
 \For{$k$ from $0$ to $s-1$}{
	Calculate assignment problem coefficients:  $\bm{a}_i^{k+1} (x)=(1- {(\nabla \text{J}_{\text{conc}})}_i(x) -\sum_{y\notin V^k_i} w(x,y))$\;
	Initialize prices $\bm{p}=\bm{0}$, incentives $\bm{t}=\bm{0}$,  $\epsilon=\epsilon_0$\;
	\While{$\epsilon\geq \bar{\epsilon}$}{
		Run Algorithm \ref{alg:fasotr} (Upper Bound Auction): $(\bm{\mathcal{X}}_{\textrm{out1}}, \bm{p}_{\textrm{out1}}, \bm{t}_{\textrm{out1}})=\textrm{ UBA }(\epsilon, \bm{B}, \bm{U}, \bm{a}^{k+1}, \bm{p},\bm{t}, \mathcal{X})$\;
		Run Algorithm \ref{alg:rasopr} (Lower Bound Auction): $(\bm{\mathcal{X}}_{\textrm{out2}}, \bm{p}_{\textrm{out2}}, \bm{t}_{\textrm{out2}})=\textrm{ LBA }(\epsilon, \bm{B}, \bm{U}, \bm{a}^{k+1}, \bm{p}_{\textrm{out1}},\bm{t}_{\textrm{out1}}, \bm{\mathcal{X}}_{\textrm{out1}})$\;
		Set $(\bm{p}, \bm{t})=( \bm{p}_{\textrm{out2}}, \bm{t}_{\textrm{out2}})$\;
		Divide $\epsilon$ by $\alpha$\;
		\If{$\epsilon<\bar{\epsilon}$}{
			Set $\bm{\mathcal{X}}^{k+1}=\bm{\mathcal{X}}_{\textrm{out2}}$\;
		}
	}

}

 \Return{($\bm{\mathcal{X}}^{s}, \bm{p}, \bm{t}$)}
 \caption{\km{Auction Dynamics} Semi-supervised learning (SSL) framework}\label{alg:advb}
\end{algorithm2e}  

\vspace{-0.25cm}

\subsection{Proposed active learning acquisition function}
\label{acq function}

In the active learning process shown in Figure \ref{fig:flowchart}, a chosen acquisition function $\mathcal{A}$ is used to select informative query points 
for which labels will be requested.
Before defining our proposed acquisition function, several variables need to be introduced. First, for each unlabeled point $x$, we calculate: 
\begin{equation}
   \label{eqn:Margin} 
   \mathcal{M}(x) =   v(x) - w(x),
\end{equation}
where $v(x) = \max_{1 \leq i \leq K}(a_i(x) - p_i + t_i + \epsilon) $ and $w(x) =\max_{1 \leq i \leq K, i \neq i^*}(a_i(x) - p_i + t_i + \epsilon).$ Here, $a_i(x)$ is the coefficient of the modified assignment problem as defined in (\ref{aucion_coefficient}), $p_i$ and $t_i$ represent the price and incentives, respectively, for class $i$, and $i^* = \arg \max_{1\leq i\leq K}(a_i(x) -p_i+t_i+\epsilon)$. As discussed in section \ref{volume_constraints}, the term $v(x)$ and $ w(x)$  represent the best and second best deal values, respectively, in the market mechanism, offered to $x$ by the end of auction method, and it is important to note that when calculating $\mathcal{M}$, one should use the final price and incentive vector returned from Algorithm \ref{alg:advb}. 
Then, the general intuition to motivate our proposed acquisition function is that there is some level of uncertainty as to the optimal class of $x$ when the value of $\mathcal{M}(x)$ is relatively small, and there is little uncertainty as to the optimal class when $ \mathcal{M}(x)$ is big.

Overall, to align with a unified framework for maximizing acquisition functions for query point selection, we define  our proposed acquisition function as follows:
\begin{equation}
\label{eqn:acquisition function} 
    \mathcal{A}(x) = 1- \mathcal{M}(x).
\end{equation}

\begin{figure*}[hbt!]
\centering 
\begin{subfigure}(a)  
\centering
\includegraphics[scale=.50]{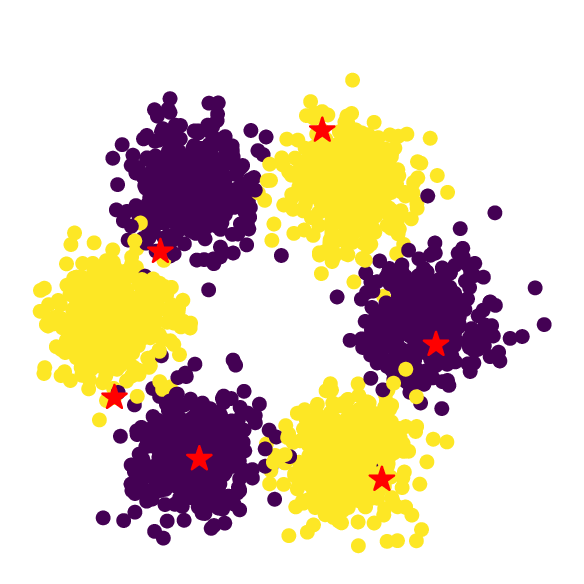} 
\end{subfigure}
\begin{subfigure}(b)  
\centering
\includegraphics[scale=.50]{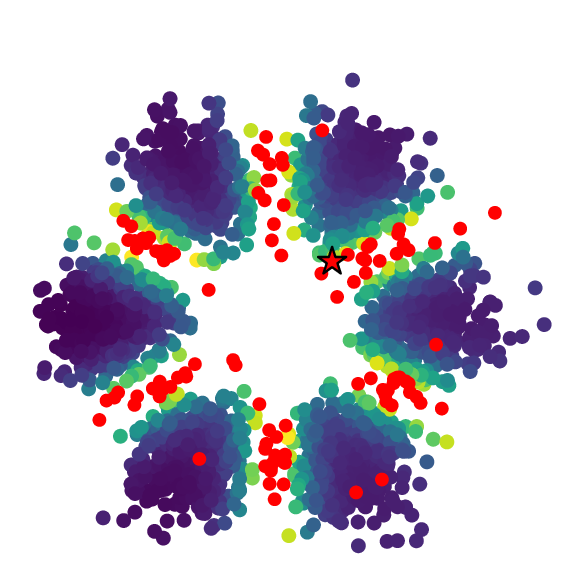} 
\label{opt}
\end{subfigure} 
 \begin{subfigure}(c)  
\centering
\includegraphics[scale=.50]{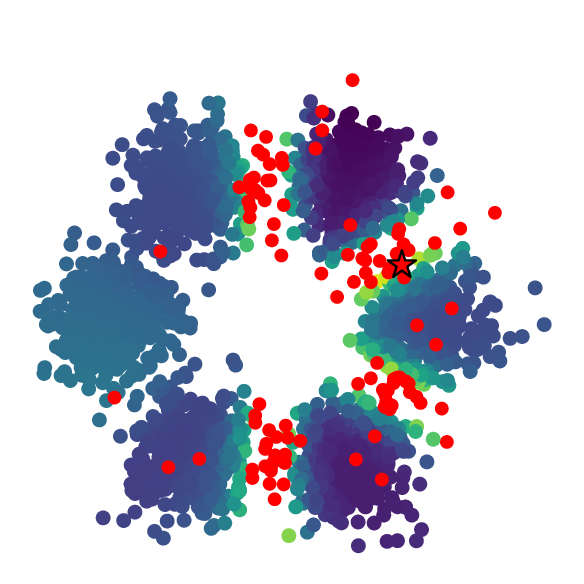} 

\end{subfigure}
\caption{ (a) Ground truth with six initial points, (b) Acquisition function value for our proposed acquisition function (\ref{eqn:acquisition function}) at Iteration 100, (c) Acquisition function value of uncertainty sampling \citep{settles_active_2012} with Laplace learning \citep{zhu2003semi} at Iteration 100.  Brighter regions of the plot indicate larger acquisition function values, and labeled points are marked in red circles; the query point for the current iteration is marked as a red star with a black outline. \km{Notice that our acquisition function (b) focuses on \emph{all} decision boundaries between oppositely labeled clusters, whereas standard uncertainty sampling (c) only focuses on a subset of these boundaries.} } 
\label{fig:Blobs_experiment}
\end{figure*}


\begin{figure*}[hbt!]
    \centering
    \includegraphics[width=0.8\linewidth]{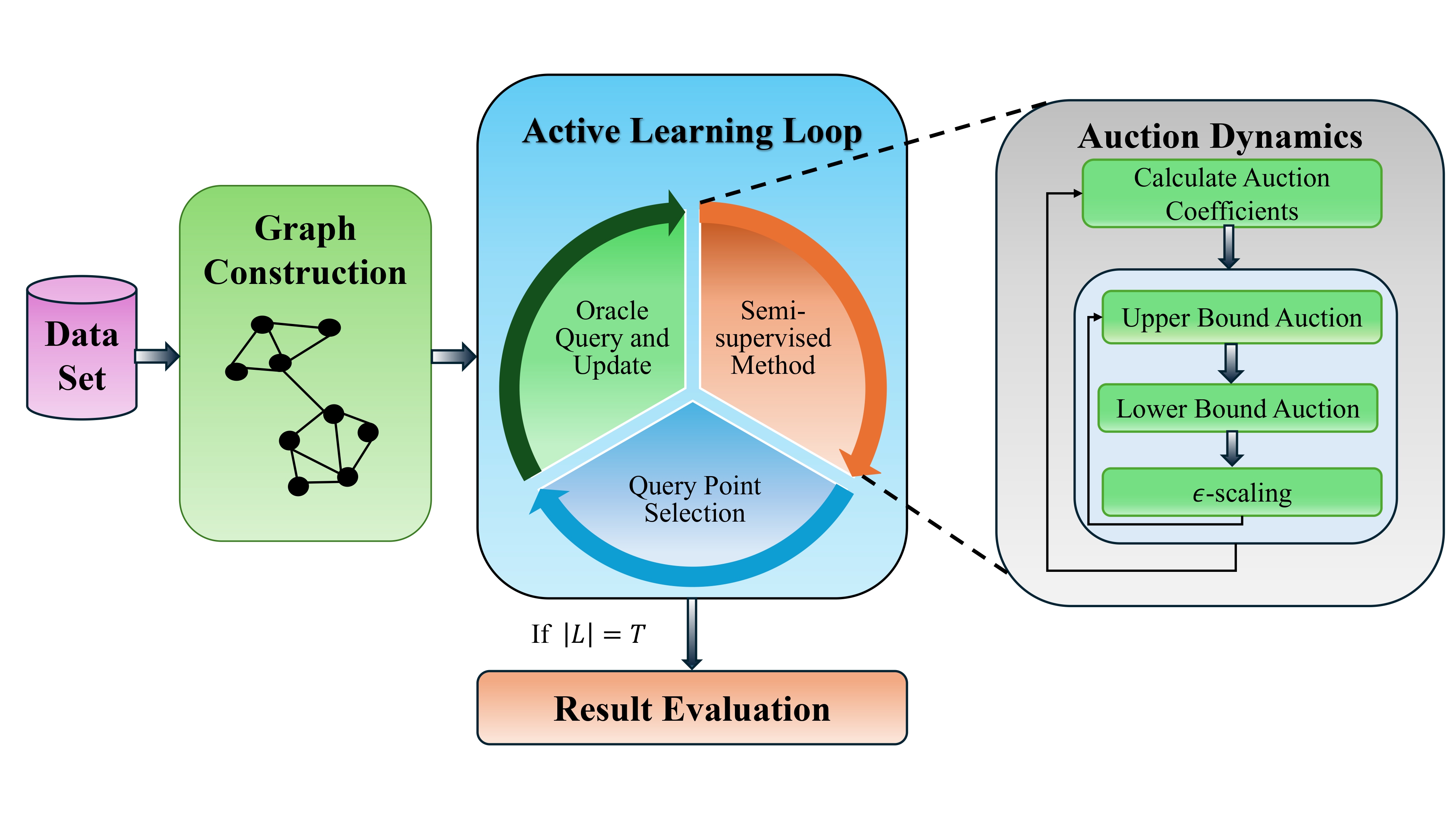}
    \caption{The flowchart of our MALADY. Green box: Similarity graph construction using a chosen similarity function (section \ref{Graph_Construction});  Blue box: Active learning process using proposed acquisition function (section \ref{acq function}); Grey box: Auction dynamics method for semi-supervised inference (section \ref{ssl}). The result is the partition of the vertex set; Red box: When $|\mathcal{L}| = T$, the accuracy of the proposed method is computed.} 
    \label{fig:malady}
    \vspace{0.5cm}
\end{figure*}

The acquisition function (\ref{eqn:acquisition function}) is a novel uncertainty sampling acquisition function \citep{settles_active_2012} where the optimal price vector $\bm{p}$, optimal incentives vector $\bm{t}$, auction coefficient $\bm{a}$ and epsilon $\epsilon$ are used in each active iteration to determine the query points. Overall, one can think of this acquisition function as prioritizing \textit{minimum margin sampling}, where the margin is defined as the difference between the best and second-best deal values offered to $x$ at the end of the auction algorithm. 
The strategy is to query a point that has the smallest margin, as a smaller margin indicates greater uncertainty in the decision. 

To gain better intuition, we run an experiment on a toy dataset in two dimensions to visualize which points are emphasized by the proposed acquisition function. The dataset consists of six Gaussian clusters, where each cluster contains 400 points and has a standard deviation of 0.25, and where the centers lie evenly spaced along the unit circle. The classification of the clusters is then assigned in an alternating pattern, as illustrated in  Figure \ref{fig:Blobs_experiment}(a). With six initially labeled points, we sequentially query 100 points for labeling via the proposed acquisition function \eqref{eqn:acquisition function}. 
\km{Heatmaps of the values for our proposed acquisition function \ref{eqn:acquisition function} and uncertainty sampling \citep{settles_active_2012} with a Laplace learning \citep{zhu2003semi} classifier are shown in panels (b) and (c), respectively. Note that brighter regions in these plots represent larger acquisition function values, helping us to visualize where the largest acquisition function values are concentrated. In our proposed active learning framework, the query point that is selected at each corresponding iteration would be the maximizer of the proposed acquisition function, representing the ``brightest'' point in the corresponding heatmap.
}

\km{
The main observation from this experiment is that \emph{the proposed acquisition function accurately prioritizes data points along all boundaries between oppositely labeled clusters}. Specifically, points at boundaries with more overlap between clusters have the highest acquisition function values. 
This contrasts with a simple uncertainty sampling \citep{settles_active_2012} acquisition function with the Laplace learning classifier \citep{zhu_combining_2003}, which in Figure \ref{fig:Blobs_experiment} (c) demonstrates that only a subset of the important boundaries between oppositely labeled clusters are emphasized in sampling.
Overall, successful active learning acquisition functions need to efficiently prioritize these decision boundary regions (referred to as ``exploitation'' in active learning \citep{miller2023poisson}). The results of this experiment suggest that the proposed acquisition function can efficiently prioritize decision boundaries modeled by regions between clusters in the corresponding similarity graph structure.  
}

Using the proposed acquisition function \eqref{eqn:acquisition function}, we turn to define our multiclass active learning method, MALADY.

\begin{algorithm2e}[h]
\SetAlgoLined
\SetKwInOut{Input}{Input}
\Input{Vertex set ${\mathcal{X}}$, weight matrix $W$, initial partition $\bm{\mathcal{X}}$, lower bound $\bm{B}$, upper bound $\bm{U}$, time step $\delta t$, number of steps $s$, auction error tolerance $\epsilon_{min}$, epsilon scaling factor $\alpha$, initial epsilon $\epsilon_0$,  \text{initial labeled set } $\mathcal{L}_0$ , 
 acquisition function $\mathcal{A}(x)$, and budget $T$. }
\KwResult{Accuracy on unlabeled set.}
 \textbf{Initialization}:  $ \mathcal{L} = \mathcal{L}_0$\;

 
	\While{$|\mathcal{L}| \leq T$}{

Run Algorithm \ref{alg:advb} (Semi-supervised method): \
 $(\bm{\mathcal{X}}^s, \bm{p}, \bm{t})=\text{ SSL } (  W,\bm{\mathcal{X}},\mathcal{\mathcal{X}}, B, U, \delta t, s, \epsilon_{min})$\;
 
		Pick $x^\ast \in \mathcal{U} = \mathcal{X} \setminus \mathcal{L} $ which has highest acquisition value  $\mathcal{A}(x)$\;
  
             $\mathcal{L} \gets \mathcal{L} \cup \{x^\ast\}$\;
		
	}
 Run Algorithm \ref{alg:advb} with labeled set $\mathcal{L}$ (Semi-supervised classification method): \
 $(\bm{\mathcal{X}}^s, \bm{p}, \bm{t})=\text{ SSL } (  W,\bm{\mathcal{X}}, \mathcal{X}, B, U, \delta t, s, \epsilon_{min})$\;
 Evaluate the performance of the classifier.\;
\Return{Evaluation of algorithm} 
\caption{MALADY: Multiclass Active Learning with Auction
DYnamics on Graphs}\label{alg:MALADY}

\end{algorithm2e}  

\vspace{-0.25cm}

\subsection{Multiclass Active Learning Procedure}
We now describe our proposed Multiclass Active Learning with Auction DYnamics on graphs (MALADY) procedure. Given an initially labeled set $\mathcal{L}_0$ with corresponding labels $\{y(x_i)\}_{x_i \in \mathcal{L}}$, we sequentially select query points that maximize our proposed acquisition function \eqref{eqn:acquisition function}, $\mathcal{A}(x) = 1 - \mathcal{M}(x)$. We summarize our method as follows: Given input data and an initially labeled subset $\mathcal{L}_0$, our proposed method consists of:
\vspace{-0.2cm}
\begin{enumerate}
    \item Construct a similarity graph from the data using a chosen similarity function such as (\ref{gaussian}). 
    \vspace{0.1cm}
    \item With a current labeled set $\mathcal{L}$, run the underlying semi-supervised classifier (Algorithm \ref{alg:advb}) to get the optimal price and incentive vectors. For the first iteration, $\mathcal{L} = \mathcal{L}_0$.
    \vspace{0.1cm}
    \item  Using acquisition function (\ref{eqn:acquisition function}), select the (next) query point as: $x^\ast = \argmax_{x \in \mathcal{U}} \mathcal{A}(x)$.
    \item Query for the label $y^\ast$ of the query point $x^\ast$ and update the labeled set $\mathcal{L} \gets \mathcal{L} \cup \{x^\ast\}$. 
    \item Repeat steps 2 through 4 until a stopping criterion, such as a budget limit $|\mathcal{L}|=T$, is met.
    \item With a current labeled set $\mathcal{L}$, run the underlying semi-supervised classifier (Algorithm \ref{alg:advb}) to get the partition of the vertex set, i.e., the data.
    \item Evaluate the classifier's performance by calculating the chosen evaluation metric on the unlabeled set. 
\end{enumerate}


The flowchart of MALADY is shown in Figure \ref{fig:malady}, and the algorithm is also described in detail as Algorithm 2.


\section{Results and Discussion} \label{Results}
\subsection{Data sets} \label{data sets}

For the computational experiments in this paper, we use eight benchmark datasets, including two hyperspectral datasets: Landsat \citep{landsat}, USPS \citep{usps}, Jasper Ridge \citep{jasper}, Urban \citep{urban}, COIL-20 \citep{krizhevsky2009learning}, CIFAR-10 \citep{nene1996columbia}, Opt-Digits \citep{optdigits}, and Fashion-MNIST \citep{xiao2017fashion}. The details of these data sets are provided in the supporting documentation.

\begin{figure*}[t!]
\centering 
\begin{subfigure}(a)  
\centering
\includegraphics[scale=.45]{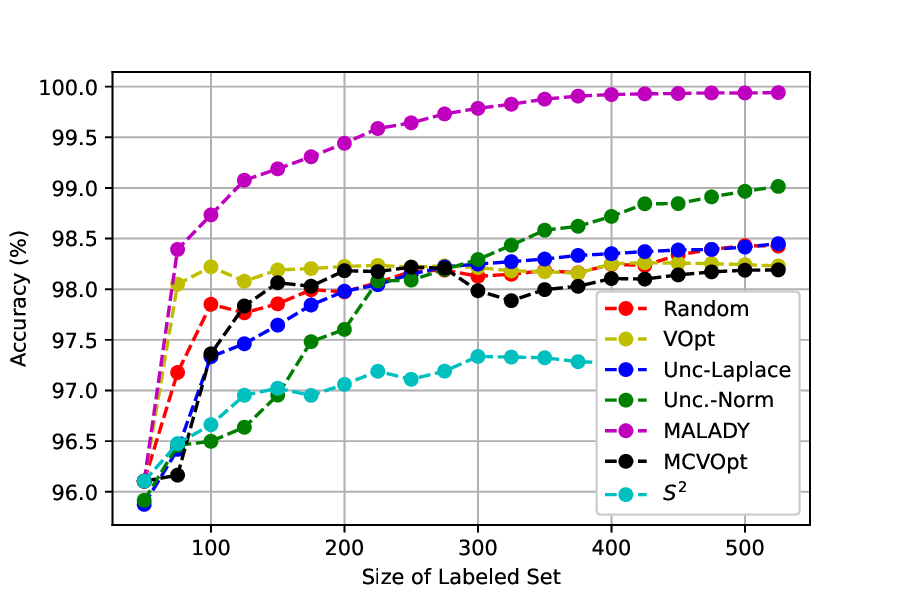} 
\label{coil}
\end{subfigure}
\begin{subfigure}(b)  
\centering
\includegraphics[scale=.45]{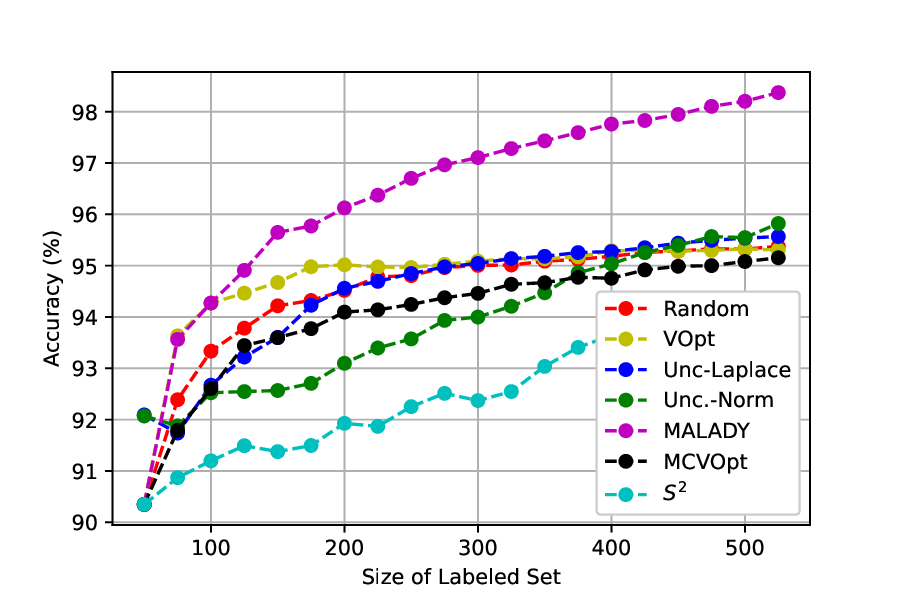} 
\label{opt}
\end{subfigure} 
 
\begin{subfigure}(c)  
\centering
\includegraphics[scale=.45]{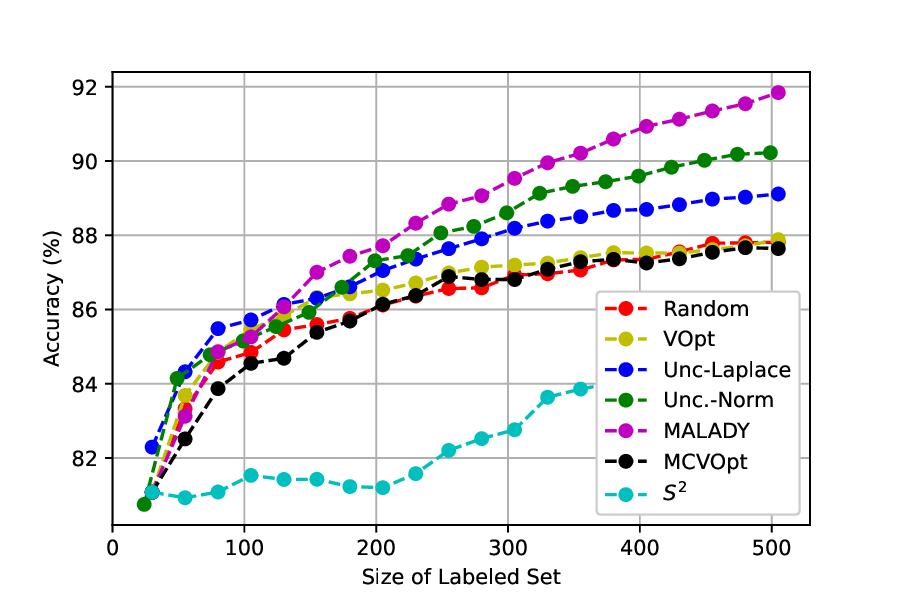} 
\label{usps}
\end{subfigure}
\begin{subfigure}(d)  
\centering
\includegraphics[scale=.45]{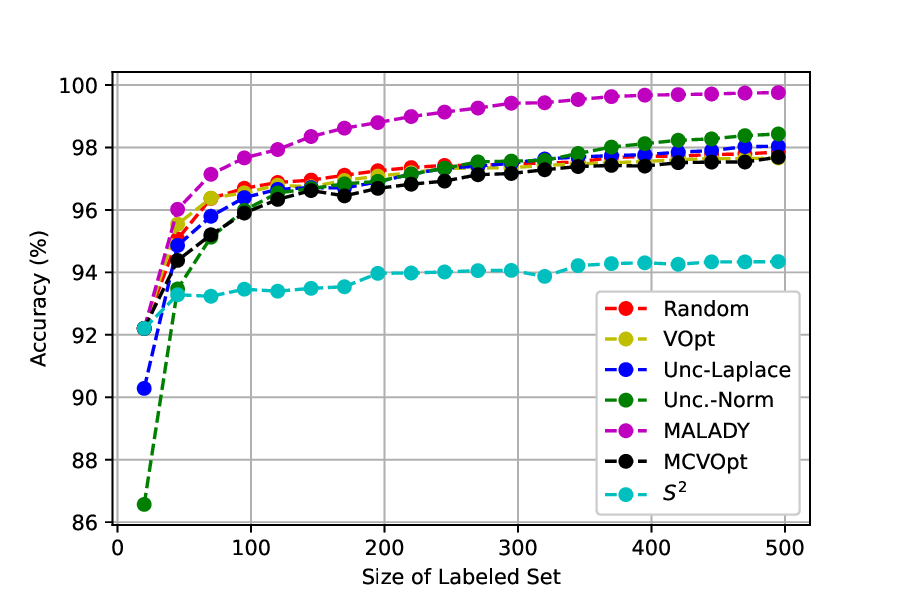} 
\label{landsat}
\end{subfigure}

\begin{subfigure}(e)  
\centering
\includegraphics[scale=.45]{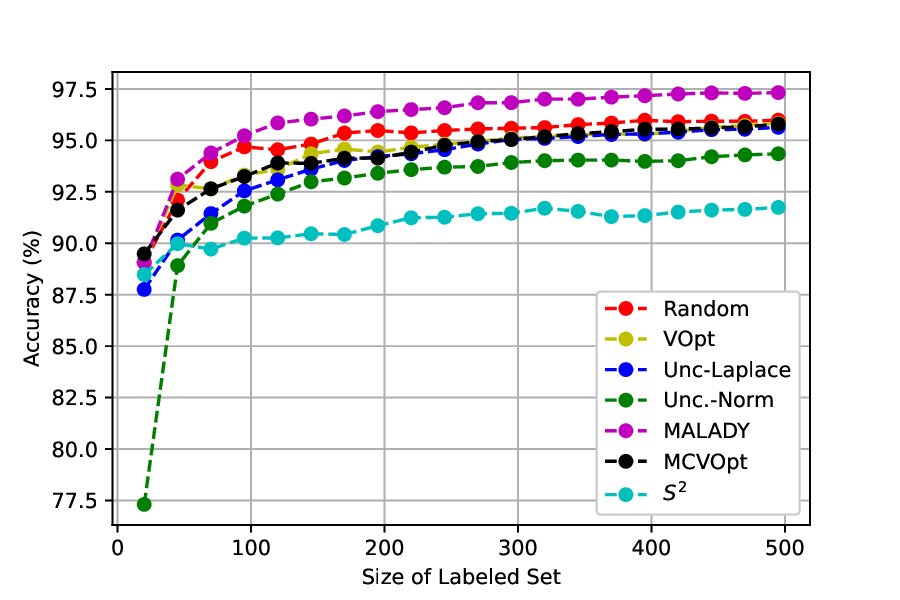} 
\label{usps}
\end{subfigure}
\begin{subfigure}(f)  
\centering
\includegraphics[scale=.45]{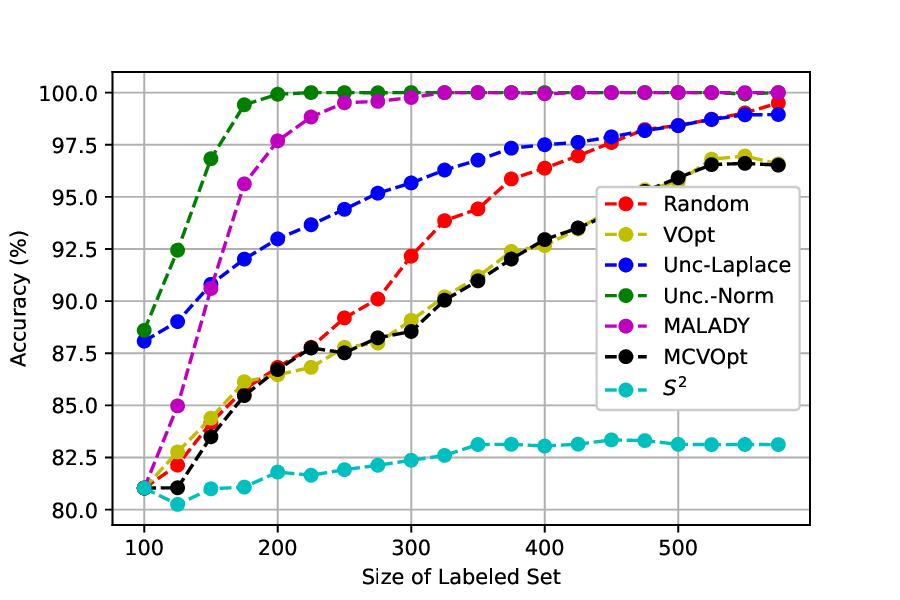} 
\label{landsat}
\end{subfigure}

\begin{subfigure}(g)  
\centering
\includegraphics[scale=.45]{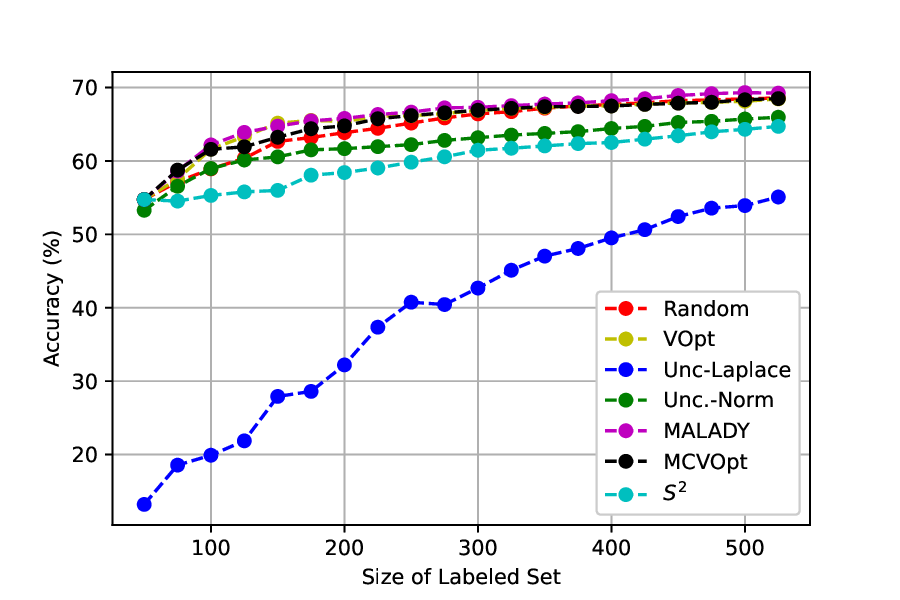} 
\label{CIFAR-10}
\end{subfigure}
\begin{subfigure}(h)  
\centering
\includegraphics[scale=.45]{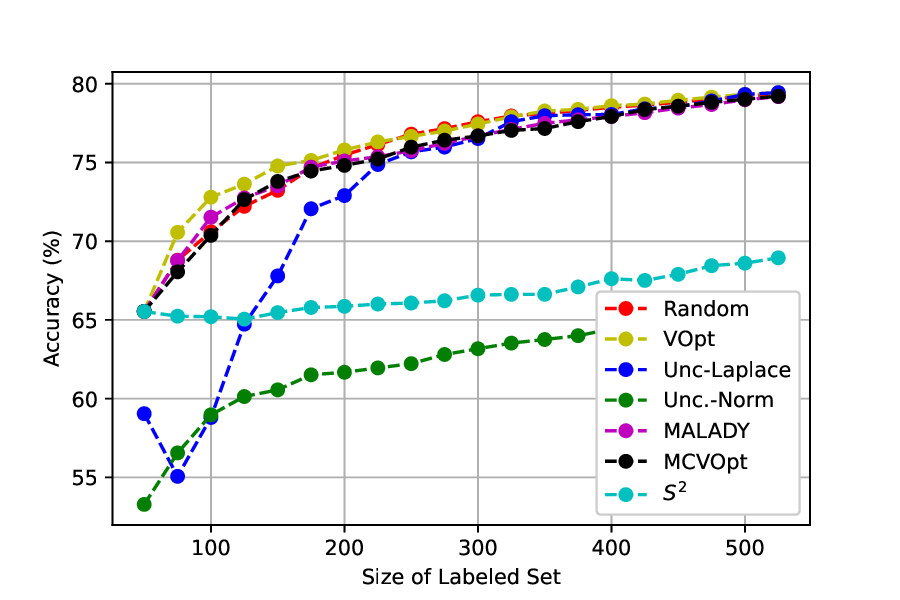} 
\label{FASHIONMNIST}
\end{subfigure}
\caption{Accuracy v.s. \# of labeled points for each dataset: (a) Opt-Digits (b) USPS (c) Satellite (d) Jasper Ridge 
 (e) Urban (f) COIL-20  (g) CIFAR-10 (h) FASHIONMNIST. \km{To allow for a more fair comparison of each acquisition function's selections for labeled sets, we report accuracy results for Random, VOpt, MCVOpt, and $S^2$ using Algorithm \ref{alg:advb} since these acquisition functions are not tied to a particular graph-based semi-supervised classifier. All other methods display accuracy results from their corresponding graph-based SSL classifiers.}} 
\label{fig:plots}
\end{figure*}

\subsection{Comparison to other methods}
In this section, we compare our proposed technique to those using other active learning acquisition functions in various graph-based SSL classifiers. The results of the experiments are shown in Figure 2. 
In particular, for all data sets, we compare our proposed method, MALADY, to the following methods: Random sampling (Random), Variance minimization criterion (VOpt)  \citep{ji_variance_2012}, Minimum norm uncertainty sampling (Unc.-Norm)\citep{miller2023poisson}, Model change active learning  (MCVOpt) \citep{miller2024model}, Uncertainty sampling (Unc.-Laplace) \citep{settles_active_2012}, and $\text{S}^2$ \citep{dasarathy2015s2}.

\km{For a fair comparison, we used Algorithm \ref{alg:advb} as the underlying SSL classifier for the Random, VOpt, MCVOpt, and $\text{S}^2$ methods since these acquisition functions are not tied to a particular, underlying graph-based semi-supervised classifier.  Minimum norm uncertainty sampling (Unc.-Norm) and Unc.-Laplace, however, are acquisition functions that at each iteration depend upon the underlying classifier's predictions to quantify uncertainty, so we report the accuracy in their originally proposed graph-based SSL classifiers; respectively, Poisson-reweighted Laplace learning \citep{calder2020properly, miller2023poisson} and Laplace learning \citep{zhu2003semi}.}

Now, we provide a brief overview of our comparison methods.
In \citep{ji_variance_2012} (VOpt method), the authors analyze the probability distribution of unlabeled vertices conditioned on the label information, which follows a multivariate normal distribution with mean corresponding to the harmonic solution across the field. The nodes are selected for querying such that the total variance of the distribution on the unlabeled data, as well as the expected prediction, is minimized.  
\km{The MCVOpt (\textit{Model Change plus VOpt}) acquisition function \citep{miller_spie_2022} is a heuristic combination of Model Change \citep{miller_model-change_2021, miller2022dissertation} and VOpt \citep{ji_variance_2012} acquisition functions. The Model Change \citep{miller_model-change_2021} acquisition function computes the change in a soft-constraint Laplace learning classifier that a currently unlabeled point could induce as a result of labeling it; this is approximated with a spectral truncation (low-rank approximation) of the graph Laplacian to improve efficiency. 
}

In minimum norm uncertainty sampling \citep{miller2023poisson}, the authors design an acquisition function that measures uncertainty in the Poisson reweighed Laplace learning algorithm (PWLL). In particular, the authors control the exploration versus exploitation tradeoff in the active learning process by introducing a diagonal perturbation in PWLL which produces exponential localization of solutions. The uncertainty sampling-based method \citep{settles_active_2012} queries points near decision boundaries that are most uncertain. Common uncertainty measures include least confidence, smallest margin, and entropy. For comparison, we use the smallest margin acquisition function with Laplace learning as an underlying classifier, which prioritizes points near the decision boundaries. 
\km{It should be noted that our proposed acquisition function for MALADY can be thought of as a form of uncertainty sampling where the quantification of uncertainty is derived from the auction dynamics graph-based classifier (Algorithm \ref{alg:advb}).}
Finally, the $\text{S}^2$ \citep{dasarathy2015s2} algorithm queries the label of the vertex that bisects the shortest-shortest path between any pair of oppositely labeled vertices.

For the implementation of MALADY, VOpt, MCVOpt, Random, and Unc.-Laplace, we utilized the GraphLearning Python package \citep{calder2020poisson} which contains both graph learning and active learning frameworks. For the implementation of Unc.-Norm \citep{miller2023poisson} and $S^2$, we utilize the code obtained from the author's repository.
\subsection{Hyperparameters Selection}
 
\begin{figure*}[t!]
\centering 
\begin{subfigure}(a)  
\centering
\includegraphics[scale=.45]{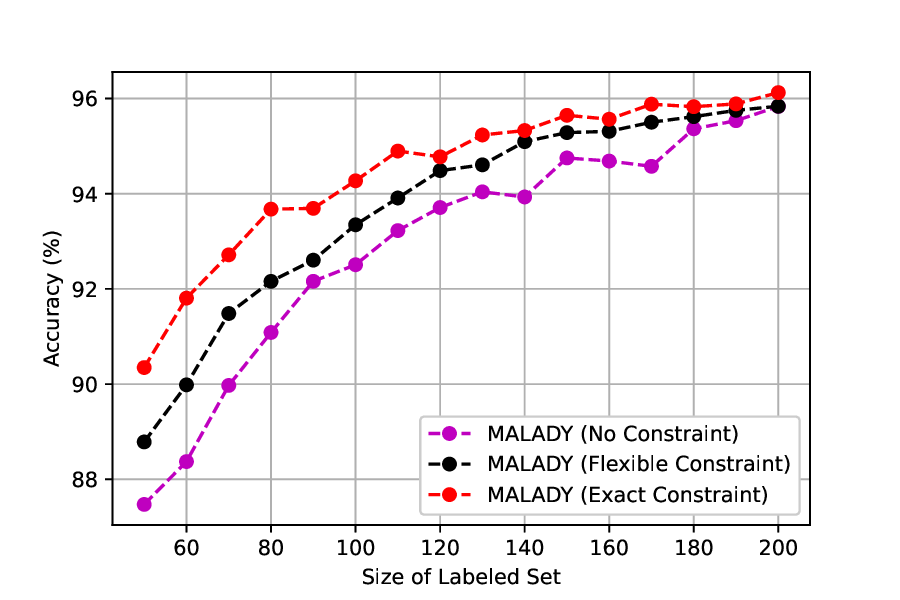} 
\label{AS_volume_bounds}
\end{subfigure}
\begin{subfigure}(b)  
\centering
\includegraphics[scale=.45]{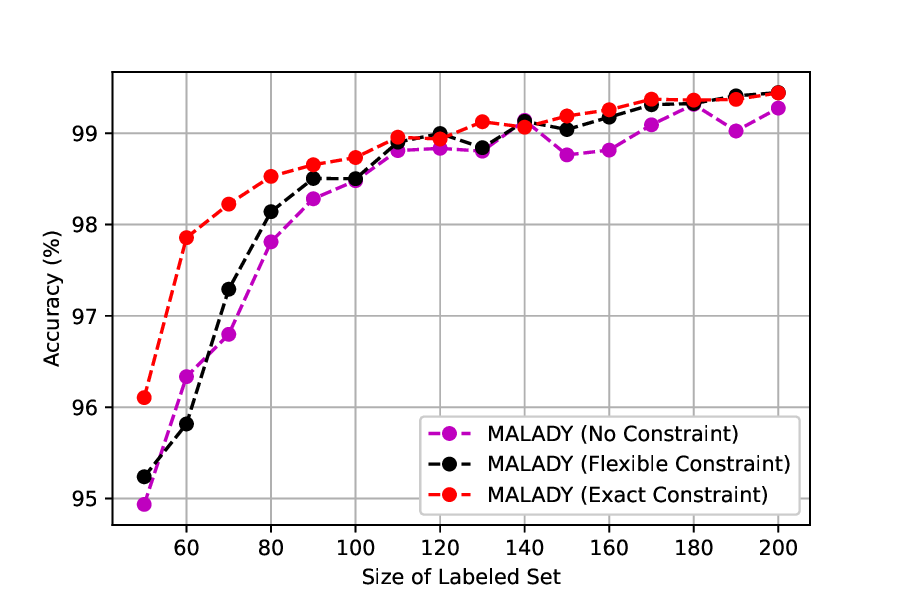} 
\label{opt}
\end{subfigure} 
 \caption{MALADY under various class size constraints. The red curve represents the exact class size constraint, the black curve represents the flexible constraint, and the pink curve represents no class size constraint. (a): USPS data set (b): Opt-Digits data set.  } 
\label{class_size_constraint}
\end{figure*}

\begin{figure*}[t!]
\centering 
\begin{subfigure}(a)  
\centering
\includegraphics[scale=.45]{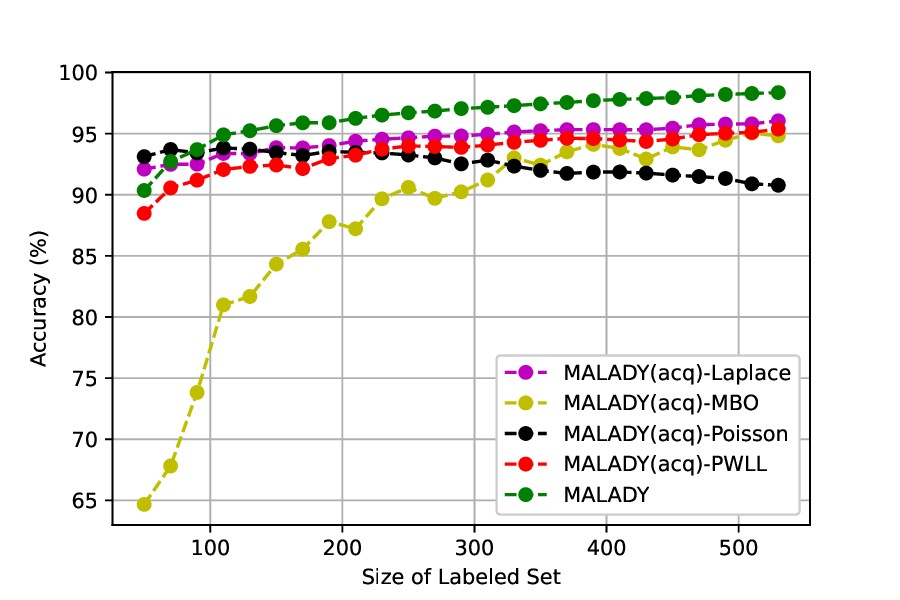} 
\label{coil}
\end{subfigure}
\begin{subfigure}(b)  
\centering
\includegraphics[scale=.45]{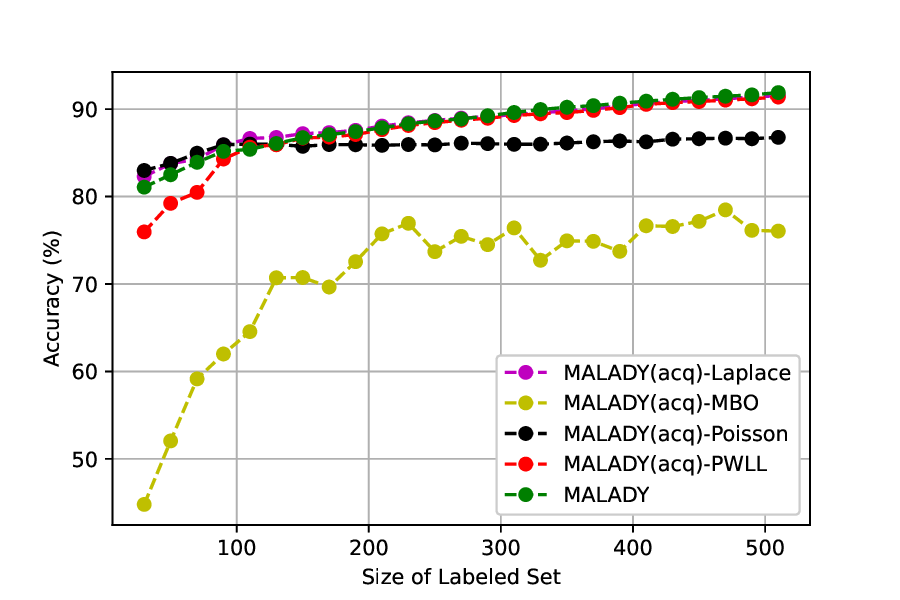} 
\label{opt}
\end{subfigure} 
 \caption{Performance of different SSL classifiers on labeled set sampled using our MALADY framework. (a): USPS data set (b): Landsat data set.  } 
\label{AS_SSL_classifier}
\end{figure*}

In this section, we outline the selected hyperparameters for our experiments. In particular, the number of nearest neighbors, $N_n$, for the weight matrix is a tunable parameter that can be adjusted to control the sparsity of the graph. 
\km{For the computational experiments, we set the number of steps $s =100$, auction error tolerance $\epsilon_{min}=1e^{-6}$, epsilon scaling factor $\alpha =4$, and the initial epsilon $\epsilon_0 = 1e^{-7}$. }
We also incorporate class size constraints for the classification, which are controlled by the upper bound $U$ and lower bound $B$. In our computational experiments, we use exact class size constraints. 
\km{See the Supplementary Information file for a comprehensive outline of all parameters used in the experiments, along with a discussion of their respective roles. }
\gokul{
\subsection{Ablation Studies}
In order to evaluate the effectiveness and robustness of our method, we perform ablation studies on 
the role of the class size constraint, the initial labeled set, and the proposed acquisition function. Firstly, to examine the performance of our method with various class size constraints, we run the experiments in two datasets using various bounds. In particular, we test USPS and Opt-Digits with various class size constraints and show the result in Figure (\ref{class_size_constraint}). We observe that our method performs better with exact constraints, especially in a low labeled rate. Second, to assess the effectiveness of the auction dynamics algorithm, we ablate on the semi-supervised learning (SSL) classifier for output prediction accuracy on unlabeled sets. Specifically, we use our MALADY framework to sample the labeled set and tested different SSL classifiers to evaluate model accuracy. The results, presented in Figure \ref{AS_SSL_classifier}, compare the performance of Multiclass MBO \citep{garcia2014multiclass}, Poisson learning \citep{calder2020poisson}, Laplace learning \citep{zhu2003semi}, and Poisson-reweighted Laplace learning (PWLL) \citep{calder2020properly}. The results in Figure \ref{AS_SSL_classifier} show that our proposed acquisition function in MALADY achieves optimal performance when paired with the auction dynamics algorithm (Algorithm \ref{alg:advb}) as the underlying SSL classifier. We also conduct additional ablation studies to understand how our framework performs with varying numbers of initially labeled samples and different distributions. Due to page limitations, however, we provide this ablation study's results in the Supporting Information file.
}

\vspace{-0.75cm}

\vspace{0.75cm}

\subsection{Performance and Discussion}
Information about the data sets used for our computational experiments is detailed in Supporting Information file. For each data set, we use five initial labels per class; thus, the initial labeled set has a total of $5K$ labeled points, where $K$ is the number of classes. We sequentially query 500 additional data elements to label. Moreover, for the hyperspectral data, we use the cosine similarity function (\ref{cosine}) for graph weight computation. For others, we use the Gaussian weight function (\ref{gaussian}).

In Figure \ref{fig:plots}, we show the accuracy performance of each acquisition function averaged over 10 trials; for all data sets, accuracy was the main evaluation metric.  From the results, one can see that our MALADY method has significantly outperformed the comparison methods in all data sets except COIL-20. For the COIL-20 data set, our MALADY algorithm performs similarly to the best-performing method Unc.-Norm \citep{miller2023poisson}, which uses a different underlying graph-based classifier than our model. 

\textcolor{black}{The significant difference in performance of our proposed MALADY method for active learning over other uncertainty-based acquisition functions (e.g., Unc.-Norm and Unc.-Laplace) warrants some further discussion. 
The success of uncertainty sampling active learning methods most often depends on the capability of the underlying classifier to use small amounts of labeled data to produce reasonable decision boundaries that can be highlighted by the corresponding acquisition function.  
Recall that our proposed acquisition function utilizes the auction values (i.e., coefficients, prices, and incentives) computed by the auction dynamics classifier (Algorithm \ref{alg:advb}) to identify the boundaries between clusters containing oppositely labeled nodes. Both the extensive empirical results (Figure \ref{fig:plots}) and the intuitive toy example visualized (Figure \ref{fig:Blobs_experiment}) suggest that these auction values are quite useful for accurately modeling the geometry of the clusters, especially these corresponding decision boundaries even with low amounts of labeled data. While a rigorous theoretical analysis of the mechanism of Algorithm \ref{alg:advb} that influences such accurate modeling of clustering geometry is outside the scope of the current work, we humbly suggest that this graph-based auction dynamics procedure is a key distinguishing component of MALADY that allows for superior uncertainty sampling performance.}

\section{Conclusion} \label{Conclusion}

 In this paper, we introduce the Multiclass Active Learning with Auction DYnamics on Graphs (MALADY) method which integrates active learning with auction dynamics techniques for semi-supervised learning and classification and ensures exploitation in the active learning process. The proposed method allows one to obtain accurate results even in the case of very small labeled sets, a common scenario for many applications since it can be costly in time and money to obtain labeled data in many applications. 
 The proposed method leverages a novel acquisition function derived from the dual variables in an auction algorithm. Lastly, the proposed algorithm also allows one to incorporate class size constraints for the classification task, which improves accuracy even further. Overall, the proposed procedure is a powerful approach for an important machine learning task.




\vspace{-0.1cm}

 \section*{Acknowledgements}

This work is supported by NSF grant DMS-2052983.



\bibliographystyle{IEEEtran}
\bibliography{IEEEabrv,main}

 \appendix

 In this section, we provide the algorithm for upper and lower bound auctions (Algorithms \ref{alg:rasopr} and \ref{alg:fasotr}, respectively) and the membership auction (Algorithm \ref{alg:faso}). The upper and lower bound auction algorithms are important for integrating the class size information. In algorithm \ref{alg:advb}, once we calculate the assignment problem coefficients, we pass them onto the upper bound auction algorithm (\ref{alg:rasopr}) to satisfy the upper bound constraint, and the result is fed into the lower bound algorithm (\ref{alg:fasotr}) to satisfy the lower bound constraint.

\begin{algorithm2e}[h]
\SetAlgoLined
\SetKwInOut{Input}{Input}
\Input{$\epsilon>0$, bounds $\bm{B}, \bm{U}$, coefficients $\bm{a}$, initial prices $\bm{p}^0$, initial admissible incentives $\bm{t}^0$, complete (but possibly lower infeasible) $\epsilon$-CS matching $\bm{\mathcal{X}^0}$ }
 \textbf{Initialization}:  Set $\bm{d}=\bm{p}^0-\bm{t}^0$, set $\bm{\mathcal{X}}=\bm{\mathcal{X}}^0$ \;
\KwResult{complete and feasible $\epsilon$-CS matching and admissible prices and admissible incentives $(\bm{\mathcal{X}},\bm{p}, \bm{t})$.}  
 \While{there exists some $i$ with  ($|\mathcal{X}_{i}|<U_{i}$ and $d_{i}>0$) or ($|\mathcal{X}_{i}|<B_{i}$)}{
 	\For{each $i^*$ with  ($|\mathcal{X}_{i^*}|<U_{i^*}$ and $d_{i^*}>0$) or ($|\mathcal{X}_{i^*}|<B_{i^*}$)}{
		\For{each $x\notin \mathcal{X}_{i^*}$}{
			Let $j$ be $x$'s current phase\;
			Calculate $\Delta(x)=(a_{j}(x)-d_j)-(a_{i^*}(x)-d_{i^*})$\;
		}
		\While{($|\mathcal{X}_{i^*}|<U_{i^*}$ and $d_{i^*}>0$) or ($|\mathcal{X}_{i^*}|<B_{i^*}$)}{
			Find $x\in \argmin_{y\notin \mathcal{X}_{i^*}} \Delta(y)$\;
			\eIf{$|\mathcal{X}_{i^*}|<B_{i^*}$}{
				Remove $x$ from its current phase and add $x$ to $\mathcal{X}_{i^*}$\;
				\If{$|\mathcal{X}_{i^*}|=B_{i^*}$ and $\Delta(x)\geq 0$}{
					Subtract $\Delta(x)+\epsilon$ from $d_{i^*}$\;
				}
			}{
				\eIf{$\Delta(x)+\epsilon\geq d_{i^*}$}{
					Set $d_{i^*}=0$\;
				}{
					Remove $x$ from its current phase and add $x$ to $\mathcal{X}_{i^*}$\;
					\If{$|\mathcal{X}_{i^*}|=U_{i^*}$ and $\Delta(x)\geq 0$}{
						Subtract $\Delta(x)+\epsilon$ from $d_{i^*}$\;
					}
				}
			
			}
		}
	}	
}
Set $\bm{p}=\max(\bm{d},\bm{0})$, set $\bm{t}=\max(-\bm{d},\bm{0})$\;
 \Return{$(\bm{\mathcal{X}},\bm{p}, \bm{t})$}
 \caption{ Lower Bound Auction (LBA)\citep{jacobs2018auction}} \label{alg:rasopr}
\end{algorithm2e}

\begin{algorithm2e}[h]
\SetAlgoLined
\SetKwInOut{Input}{Input}
\Input{$\epsilon>0$,  bounds $\bm{B}, \bm{U}$, coefficients $\bm{a}$,  initial prices $\bm{p}^0$ and incentives $\bm{t}^0$, data $x\in \mathcal{X}$}
\KwResult{Prices $\bm{p}$, incentives $\bm{t}$, and complete $\epsilon$-CS matching $\bm{\mathcal{X}}$ satisfying upper bounds.  }
 \textbf{Initialization}: All $x$ unassigned, $\bm{d}\hspace{-0.05cm}=\hspace{-0.05cm}\bm{p}^0\hspace{-0.05cm}-\hspace{-0.05cm}\bm{t}^0$, $\bm{\mathcal{X}}\hspace{-0.05cm}=\hspace{-0.05cm}\varnothing$
 \While{some $x$ is marked as unassigned}{
 	\For{each unassigned $x\in \mathcal{X}$}{
		Calculate $i_{cs}(x,\bm{p})$; choose $i^*\in i_{cs}(x,\bm{d})$\;
		$b(x)=d_{i^*}+\epsilon+ (a_{i^*}(x)-d_{i^*})-(a_{i_{\textrm{next}}}(x)-d_{i_{\textrm{next}}})$\;
		\uIf{$|\mathcal{X}_{i^*}|=U_{i^*}$}{
			Find $y=\argmin_{z\in \mathcal{X}_{i^*}} b(z)$\;
			Remove $y$ from $\mathcal{X}_{i^*}$ and add $x$ to $\mathcal{X}_{i^*}$\;
			Mark $y$ as unassigned, $x$ as assigned\;
			Set $d_{i^*}=\min_{z\in \mathcal{X}_{i^*}} b(z)$\;
		}\uElseIf{$|\mathcal{X}_{i}|=B_{i}$ and $d_{i}<0$}{
				Find $y=\argmin_{z\in \mathcal{X}_{i^*}} b(z)$\;
				Remove $y$ from $\mathcal{X}_{i^*}$ and add $x$ to $\mathcal{X}_{i^*}$\;
				Mark $y$ as unassigned, $x$ as assigned\;
				Set $d_{i^*}=\min(\min_{z\in \mathcal{X}_{i^*}} b(z),0)$\;
		}\Else{
			Mark $x$ as assigned and add $x$ to $\mathcal{X}_{i^*}$\;
		}	
	}
  }
Set $\bm{p}=\max(\bm{d},\bm{0})$, set $\bm{t}=\max(-\bm{d},\bm{0})$\;
 \Return{$(\bm{\mathcal{X}},\bm{p}, \bm{t})$}
 \caption{Upper Bound Auction (UBA) \citep{jacobs2018auction} }\label{alg:fasotr}
\end{algorithm2e}

 \begin{algorithm2e}[h]
\SetAlgoLined
\SetKwInOut{Input}{Input}
\Input{$\epsilon>0$,  volumes $\bm{V}$, coefficients $\bm{a}$,  initial prices $\bm{p}^0$, data $x\in \mathcal{X}$}
\KwResult{Final $\bm{p}$, complete $\epsilon$-CS matching $(\bm{\mathcal{X}},\bm{p})$.  }
 \textbf{Initialization}: For every $i\in \{1,\ldots, N\}$ mark all $x$ as unassigned, set $\bm{p}=\bm{p}^0$, set $\bm{\mathcal{X}}=\varnothing$ \;	
 \While{some $x$ is marked as unassigned}{
 	\For{each unassigned $x\in \mathcal{X}$}{
		Calculate $i_{cs}(x,\bm{p})$; choose $i^*\in i_{cs}(x,\bm{p})$\;
		$b(x)=p_{i^*}+\epsilon+ (a_{i^*}(x)-p_{i^*})-(a_{i_{\textrm{next}}}(x)-p_{i_{\textrm{next}}})$\;
		\eIf{$|\mathcal{X}_{i^*}|=V_{i^*}$}{
			Find $y=\argmin_{z\in \mathcal{X}_{i^*}} b(z)$\;
			Remove $y$ from $\mathcal{X}_{i^*}$ and add $x$ to $\mathcal{X}_{i^*}$\;
			Mark $y$ as unassigned, $x$ as assigned\;
			Set $p_{i^*}=\min_{z\in \mathcal{X}_{i^*}} b(z)$\;
		}{
			Mark $x$ as assigned and add $x$ to $\mathcal{X}_{i^*}$\;
			\If{$|\mathcal{X}_{i^*}|=V_{i}$}{
				Set $p_{i^*}=\min_{z\in \mathcal{X}_{i^*}} b(z)$\;
			}
		}	
	}
  }
 
 \Return{$(\bm{\mathcal{X}},\bm{p})$}
 \caption{Membership Auction \citep{bertsekas1989auction, jacobs2018auction}}\label{alg:faso}
\end{algorithm2e}

\end{document}